\documentclass[10pt,twocolumn,letterpaper]{article}
\usepackage[OT1]{fontenc}
\PassOptionsToPackage{table}{xcolor}
\usepackage[pagenumbers]{cvpr}
\usepackage{tabularx,array,multirow,pifont,tcolorbox,algorithm}

\definecolor{cvprblue}{rgb}{0.21,0.49,0.74}
\usepackage[breaklinks,colorlinks,allcolors=cvprblue,hyperfootnotes=false]{hyperref}

\definecolor{dominoBlue}{HTML}{405E83}

\newtcolorbox{dominoprompt}[2]{
    width=\linewidth,
    colback=#1!3!white,
    colframe=#1!45!white,
    colbacktitle=#1!12!white,
    coltitle=#1!70!black,
    fonttitle=\small\bfseries,
    fontupper=\small,
    title={#2},
    boxrule=0.5pt,
    arc=1.5mm,
    left=2mm, right=2mm,
    top=1.5mm, bottom=1.5mm,
    before skip=6pt, after skip=6pt
}

\DeclareFontFamily{TU}{dominoubuntu}{}
\DeclareFontShape{TU}{dominoubuntu}{m}{n}
  {<-> "[assets/ubuntu/Ubuntu-Regular.ttf]:mapping=tex-text"}{}
\DeclareFontShape{TU}{dominoubuntu}{b}{n}
  {<-> "[assets/ubuntu/Ubuntu-Medium.ttf]:mapping=tex-text"}{}

\newcommand{\UbuntuTitle}{\usefont{TU}{dominoubuntu}{m}{n}}
\newcommand{\DominoMainSectionFace}{\usefont{TU}{dominoubuntu}{b}{n}}

\catcode`’=\active
\protected\def’{\textquoteright}

\definecolor{TitleInk}{HTML}{10217C}
\definecolor{TitlePanel}{HTML}{F3F5FA}
\hypersetup{bookmarksnumbered=true,pdfdisplaydoctitle=true}

\let\RestoreCVPRSectionFormat\cvprsection
\newcommand{\UseMainSectionFont}{%
  \patchcmd{\cvprsection}
    {\large\bf}
    {\large\DominoMainSectionFace}
    {}
    {\PackageError{domino}{Could not apply the main section font}
       {Check the section definition in cvpr.sty.}}%
}
\newcommand{\RestoreCVPRSectionFont}{%
  \let\cvprsection\RestoreCVPRSectionFormat
}

\newcommand{\DominoAuthor}[2]{%
  \mbox{#1\textsuperscript{#2}}%
}

\renewcommand{\maketitle}{%
  \begin{tcolorbox}[
    colback=TitlePanel,
    colframe=TitlePanel,
    boxrule=0pt,
    arc=9pt,
    outer arc=9pt,
    boxsep=0pt,
    left=12pt,
    right=12pt,
    top=11pt,
    bottom=9pt,
    before skip=0pt,
    after skip=8pt
  ]
    \centering

    {\UbuntuTitle\fontsize{14.1}{16.2}\selectfont
      \color{TitleInk}\PaperTitleLines{\\}\par}

    \vspace{8pt}

    {\UbuntuTitle\fontsize{9.3}{11}\selectfont
      \DominoAuthor{Jeahun Sung}{1,*}\quad
      \DominoAuthor{Seungho Choi}{1,*}\quad
      \DominoAuthor{Taekyung Kim}{2,\textdagger}\quad
      \DominoAuthor{Jihyong Oh}{1,\textdagger}\par}

    \vspace{1pt}

    \noindent
    \begin{minipage}[c]{.19\linewidth}
      \centering
      \includegraphics[width=80pt]{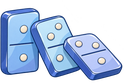}
    \end{minipage}\hfill
    \begin{minipage}[c]{.56\linewidth}
      \centering
      \fontsize{8.7}{10.3}\selectfont
      \begin{tabular}{@{}c@{\hspace{13pt}}c@{}}
        \textsuperscript{1}\,CMLab, Chung-Ang University &
        \textsuperscript{2}\,NAVER AI Lab\\[2pt]
        {\fontsize{7.3}{8.7}\selectfont
          (\{jhseong, choiseungho1019, jihyongoh\}@cau.ac.kr)} &
        {\fontsize{7.3}{8.7}\selectfont
          (taekyung.k@navercorp.com)}
      \end{tabular}
    \end{minipage}\hfill
    \begin{minipage}[c]{.19\linewidth}
      \centering
      \includegraphics[width=80pt]{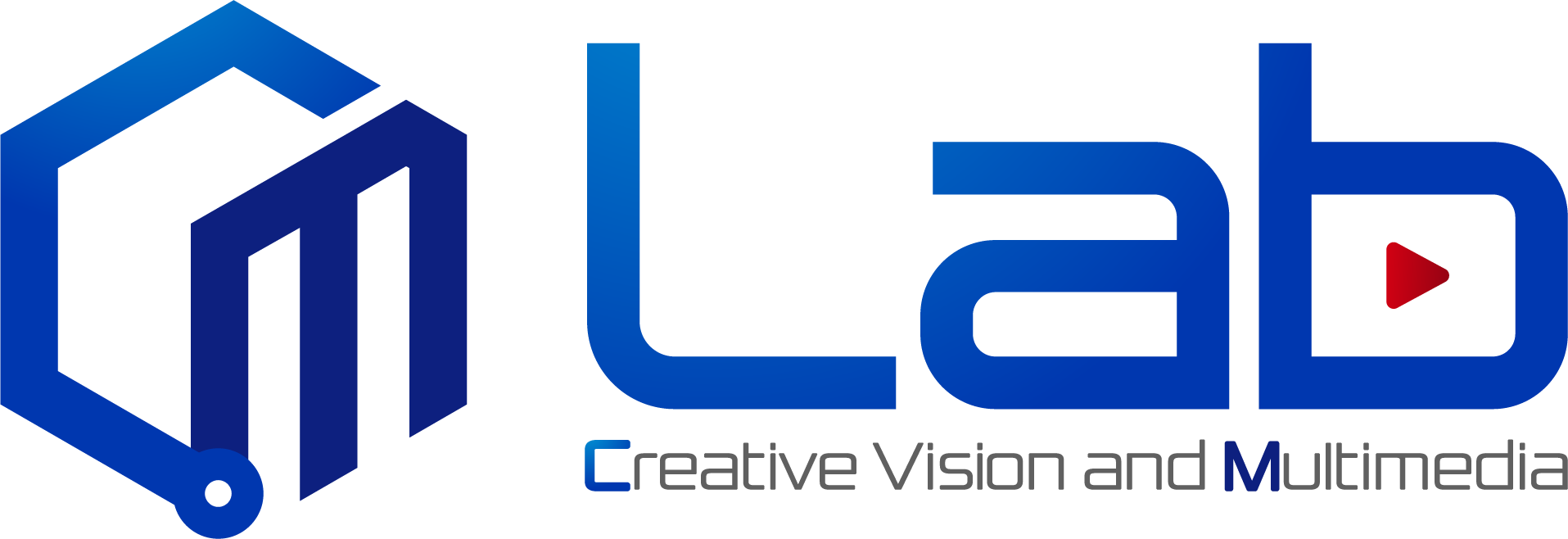}\\[6pt]
      \includegraphics[width=76pt]{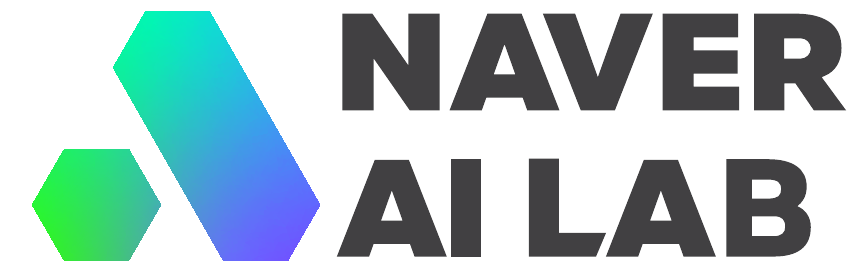}
    \end{minipage}\par

    \vspace{4pt}
    {\centering
      \fontsize{8}{9.5}\selectfont
      \color{TitleInk}%
      \ifdefvoid{\ProjectPageURL}
        {\def\DominoURL{https://cmlab-korea.github.io/DOMINO}}
        {\let\DominoURL\ProjectPageURL}%
      \hypersetup{urlcolor=TitleInk}%
      \urlstyle{tt}%
      \textbf{Project page:}\enspace
      \expandafter\url\expandafter{\DominoURL}\par}

  \end{tcolorbox}%
}

\AddToHook{bfseries}{\DominoMainSectionFace}
\renewcommand*{\elvbf}{\fontsize{11}{13.2}\DominoMainSectionFace}

\DeclareCaptionFont{dominomedium}{\DominoMainSectionFace}

\newcommand{\PaperTitleLines}[1]{DOMINO: A Cause-and-Effect-Aware Graph-based Agent and#1Real-World Benchmark for Physically Plausible Multi-Turn Image Editing}
\newcommand{\PaperTitle}{\PaperTitleLines{\space}}
\title{\PaperTitle}
\author{Jeahun Sung \and Seongho Choi \and Taekyung Kim \and Jihyong Oh}
\hypersetup{pdftitle={\PaperTitle},pdfauthor={Jeahun Sung, Seongho Choi, Taekyung Kim, Jihyong Oh}}

\providecommand{\cmark}{\textcolor{green!50!black}{\ding{51}}}
\providecommand{\xmark}{\textcolor{red!80!black}{\ding{55}}}

\definecolor{checkgreen}{RGB}{0,145,65}
\definecolor{crossred}{RGB}{210,45,45}

\newcommand{\modelname}{DOMINO}
\newcommand{\benchmarkname}{\textcolor{black}{DOMINO-Bench}}

\definecolor{bluejh}{RGB}{0, 0, 0}
\newcommand{\jh}[1]{\textcolor{bluejh}{#1}}
\definecolor{redsh}{RGB}{255, 0, 0}

\newcommand{\ProjectPageURL}{}
\begin{document}
\twocolumn[{%
\begin{minipage}{\textwidth}
\maketitle
\begin{center}
\begin{minipage}{\linewidth}
\centering
\vspace{-0.25cm}
\includegraphics[width=.97\textwidth]{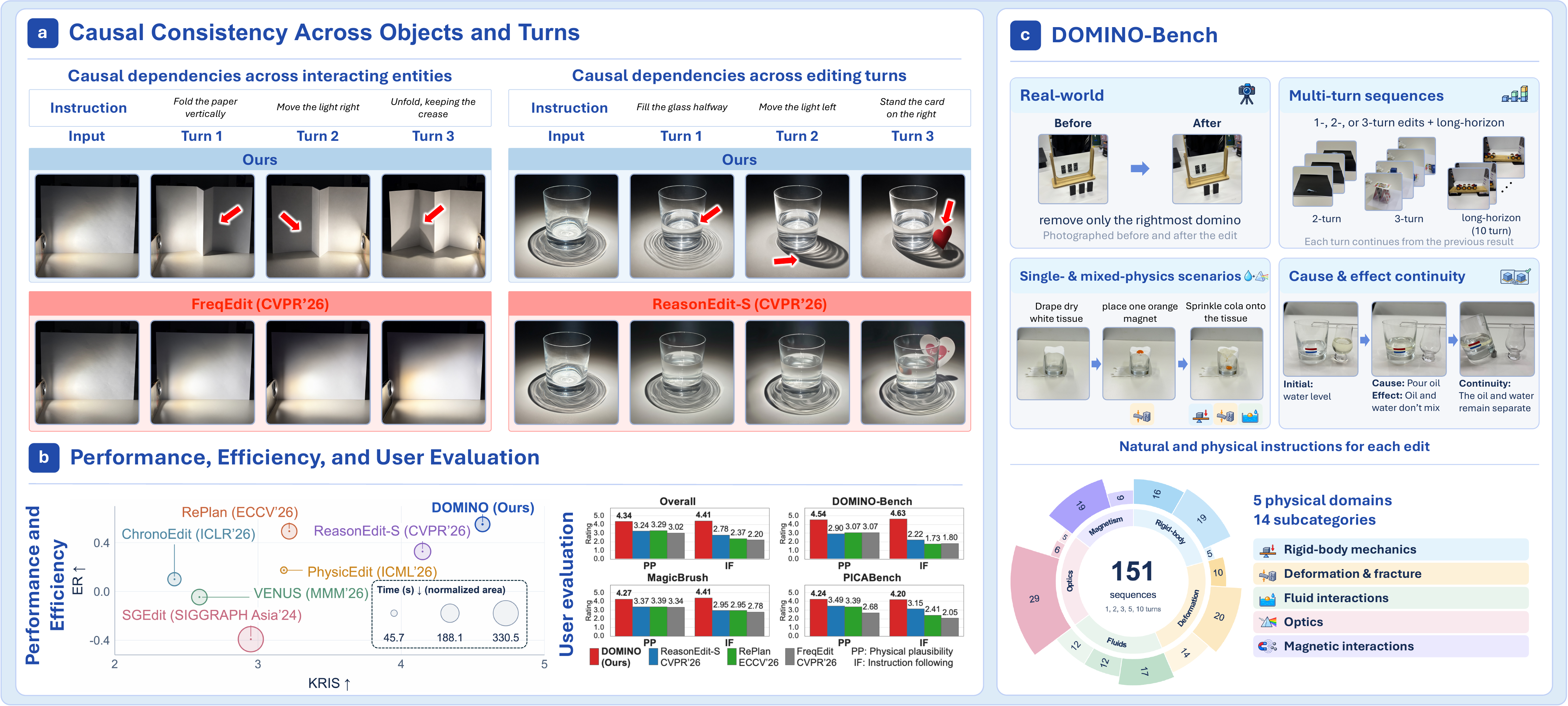}
    \captionsetup{hypcap=false}
\captionof{figure}{
    \textbf{Overview of DOMINO and DOMINO-Bench.}
    (a) DOMINO updates physical states and preserves
    unaffected attributes across interacting entities
    and editing turns.
    (b) \textbf{KRIS Knowledge}~\cite{wu2026kris} and
    \textbf{EditReward (ER)}~\cite{wu2026editreward} assess
    consistency with real-world knowledge and editing quality,
    respectively.
    DOMINO leads both metrics; bubble area indicates runtime.
    In a separate user evaluation, DOMINO also receives
    the highest ratings for instruction following
    and physical plausibility.
    (c) DOMINO-Bench provides 151 real-world sequences
    across five physical domains and 14 subcategories,
    covering single- and mixed-physics scenarios
    and short- and long-horizon editing.
}
    \label{fig:teaser}
\end{minipage}
\end{center}
\end{minipage}\par\vspace{5pt}
\vspace{-0.2cm}
}]

\begingroup
\renewcommand{\thefootnote}{\fnsymbol{footnote}}
\footnotetext[1]{Co-first authors (equal contribution).}
\footnotetext[2]{Co-Corresponding author.}
\endgroup

\begin{abstract}
\textit{Instruction-guided image editors can perform requested edits
while overlooking their physical consequences.
In multi-turn editing, these consequences depend on states
established by earlier edits, requiring both physical updates
and state preservation.
To this end, we propose \textbf{\modelname{}}, a training-free, cause-and-effect-aware
graph-based agent for physically plausible multi-turn image editing.
Our framework consists of two stages:
cause-and-effect-aware planning (Stage~1)
and physically plausible image editing (Stage~2).
In Stage~1, Weaver builds a scene graph of physical states
and interactions to support reasoning about physical consequences.
Using this graph, Forecaster infers these consequences
and constructs an editing plan specifying which attributes
to update or preserve.
Sentinel reviews the plan for physical plausibility and conflicts
between update and preservation requirements,
providing feedback for refinement.
In Stage~2, Consequence-to-Region Attention (CRA)
translates the reviewed plan into spatial guidance
for a frozen editor. Each edited image provides the scene state for planning
the next turn.
We further construct \textbf{DOMINO-Bench}, a real-world benchmark comprising 151 editing sequences with single- and mixed-physics scenarios. It uses photographed outcomes to evaluate physical consequences and cause-and-effect continuity, testing how earlier changes affect later outcomes.
Together, the framework and benchmark connect physical
reasoning, spatial edit control, and real-world evaluation
of state consistency across successive edits.}
\end{abstract}
\clearpage

\UseMainSectionFont

\section{Introduction}
\label{sec:introduction}

Instruction-guided image editing allows users to modify images
through natural-language requests~\cite{brooks2023instructpix2pix}.
Beyond visual content creation, physically plausible editing
can support world simulation by depicting the outcomes of
physical interventions~\cite{wu2026chronoedit,zeng2025editworld}.
However, current editors~\cite{labs2025flux,wu2025qwen} can perform
the requested edit yet produce inconsistent reflections or implausible
deformations~\cite{pu2026picabench,zhao2026from}.
Addressing these physical inconsistencies requires reasoning
about \textit{cause-and-effect} relationships to infer the
physical consequences left implicit in natural instructions~\cite{pu2026picabench,wan2026physedit}.

Physically plausible multi-turn image editing requires
accounting for causal dependencies at two levels:
\textbf{(i) across interacting entities} and
\textbf{(ii) across editing turns}.
First, \textbf{accounting for causal dependencies across interacting
entities} requires distinguishing physical state updates
from attributes that should be preserved.
Changing the paper's shape also requires updating its
shadows while preserving its identity and material
(Fig.~\ref{fig:teaser}(a), left).
An editing plan must therefore identify the affected entities
and specify both physical state updates and attribute-specific
preservation requirements.
Second, \textbf{accounting for causal dependencies across editing turns}
requires reasoning from the physical conditions established
by earlier edits.
When a later turn adds the heart, the edit must preserve
the previously added water and produce a shadow consistent
with the lighting established earlier
(Fig.~\ref{fig:teaser}(a), right).
Each turn must determine which earlier physical
states to update or preserve.

To address these requirements, we propose \modelname{}
as a training-free,
\textbf{cause-and-effect-aware graph-based agent}
for physically plausible multi-turn image editing.
The method operates in two stages: planning by the agent
and image editing with a frozen editor.
In the Stage~1 planning phase, the agent constructs and reviews an editing plan
through three subagents:
\textbf{Weaver} for scene graph construction,
\textbf{Forecaster} for physical consequence inference
and planning, and
\textbf{Sentinel} for plan review.
Scene graphs provide a structured interface for editing
entities and their relations~\cite{SGEdit,VENUS}.
At each turn, the Weaver extracts a scene graph representing
the current entities, their physical states and interactions,
and the associated image regions.
The Forecaster infers physical consequences and records them
in a causal specification.
It then constructs an editing plan that represents the
intended scene with a target graph and specifies physical
state updates and preservation requirements.
The Sentinel reviews the plan for instruction following,
physical plausibility, and consistency with the causal
specification and preservation requirements.
The Forecaster uses this feedback to refine the plan.

In the Stage~2 image editing phase, the \textbf{Consequence-Guided Editor} applies
the reviewed editing plan to the current image
(Sec.~\ref{sec:attention_control}).
A global text description of the editing plan leaves the
image editor to infer where each physical state update and
preservation requirement should apply~\cite{qu2026replan}.
We therefore propose \textbf{Consequence-to-Region Attention
(CRA)} as an attention-control module within the
\textbf{Consequence-Guided Editor}.
CRA uses graph-derived consequence-to-region associations
to guide a frozen image editor~\cite{teamqwen} through
attention control~\cite{hertz2023prompttoprompt,densediffusion}.
The plan associates text clauses describing physical
state updates and attribute-specific preservation requirements
inferred during planning with regions of relevant entities
and interaction effects.
CRA encourages attention to each clause within its associated image region and discourages attention to that clause outside
the region.
Update and preservation clauses can share a region, allowing
the same region to receive both types of guidance.
These explicit associations also enable \textbf{interactive
control}, allowing users to inspect the editing plan and
adjust the associated image-region boxes.
The edited image becomes the input to the next turn, where
the scene is observed again with a new instruction.

Evaluating these capabilities requires real-world references
and sequences that capture physical interactions across turns.
PICABench~\cite{pu2026picabench} lacks physically captured
target outcomes.
PhyEditBench~\cite{guo2026phyeditbench} evaluates video-derived
physical trajectories but does not explicitly target
multi-physics interactions or cause-and-effect continuity
across distinct editing interventions.
To address these gaps, we construct \benchmarkname{},
a real-world benchmark comprising 151 editing sequences
with single- and mixed-physics scenarios.
These sequences assess cause-and-effect continuity by testing
how earlier physical changes influence later outcomes.
Each turn pairs a natural instruction for model input
with a physical instruction for evaluation.

\begin{figure*}[!t]
    \centering
    \includegraphics[width=1.00\linewidth]{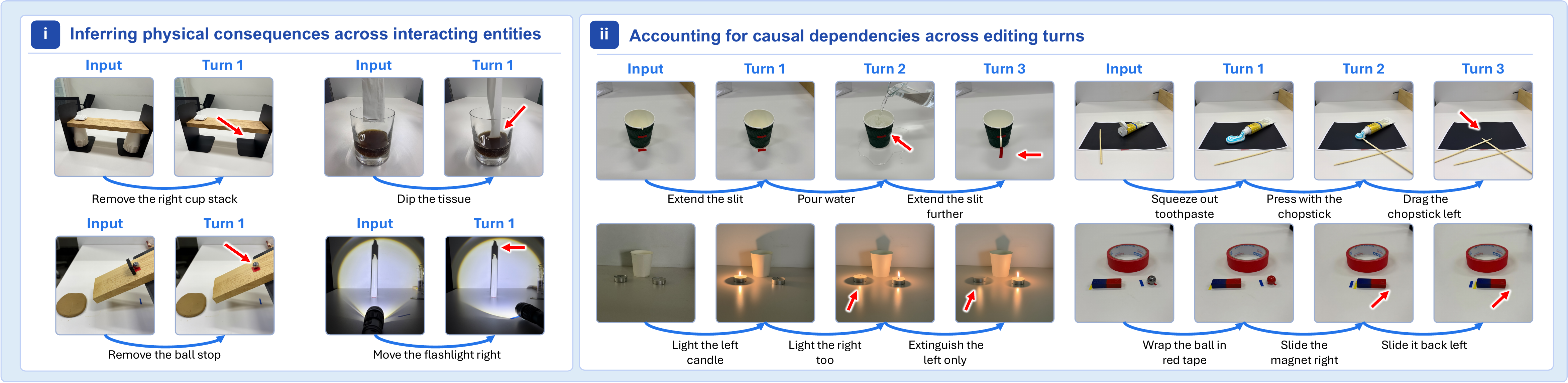}
    \caption{
        \textbf{Challenges in physically plausible multi-turn
        image editing.}
        Results from CoCoEdit~\cite{wu2026cocoedit} show two challenges:
        \textit{(i) inferring physical consequences across
        interacting entities} and
        \textit{(ii) accounting for causal dependencies
        across editing turns}.
        The board remains unchanged after support removal,
        and a previously extended slit disappears during
        the pouring edit.
        Similar failures in other editors are shown in
        Appendix~\ref{sec:supp_qualitative_results}.
    }
    \label{fig:motivation}
\end{figure*}

Our contributions are as follows:
\begin{itemize}
    \item
    We propose \modelname{}, a training-free,
    cause-and-effect-aware graph-based agent
    for physically plausible multi-turn image editing
    in two stages: planning and editing.

    \item
    \textbf{Stage~1: Cause-and-Effect-Aware Planning.}
    The agent infers physical consequences from the current
    scene and constructs and reviews editing plans using
    scene graphs, specifying physical state updates and
    attribute-specific preservation requirements.

    \item
    \textbf{Stage~2: Physically Plausible Image Editing.}
    We propose CRA within the Consequence-Guided Editor
    to associate physical state updates and attribute-specific
    preservation requirements with image regions, providing
    spatial attention guidance for a frozen image editor
    and interactive control via image-region box adjustment.

    \item
    We construct \benchmarkname{} from sequential real-world
    physical interventions to evaluate physical consequences
    across interacting entities and editing turns,
    including long-horizon physical plausibility and
    preservation of physical states unaffected by later
    interventions.
\end{itemize}
We provide a more detailed discussion of Related Work
in Appendix~\ref{sec:supp_related_work}.

\section{\modelname{}} 
\label{sec:pm} 
\subsection{Motivation}

Physically plausible multi-turn image editing poses two
related challenges: \textit{(i) inferring physical consequences
across interacting entities} and \textit{(ii) accounting for
causal dependencies across editing turns}.
A natural instruction describes the requested action, but
does not necessarily specify all the changes required for a
physically plausible result~\cite{pu2026picabench}.
For example, removing the right cup stack requires updating
the board's position, yet the edited board remains unchanged
(Fig.~\ref{fig:motivation}(i)).
Across turns, pouring water into a cup must preserve
a slit extended in an earlier turn and account for its
effect on water flow.
However, the slit disappears during the pouring edit
(Fig.~\ref{fig:motivation}(ii)).
The required changes therefore depend on the current physical
states, including those established by earlier edits.
Reasoning about these \textit{cause-and-effect} relationships
requires identifying the affected entities and determining
which attributes to update and which to preserve.

Motivated by these observations, we use scene graphs, which
provide a structured interface for editing entities and their
relations~\cite{SGEdit,VENUS}.
Representing physical states and interactions explicitly
supports reviewing planned changes against inferred physical
consequences and preservation requirements.
However, describing the plan only in a global text prompt
leaves the image editor to infer where each condition
should apply~\cite{qu2026replan}.
Attention-control methods regulate the spatial influence
of text conditions~\cite{hertz2023prompttoprompt,densediffusion}.
For physically plausible editing, this guidance must associate
the inferred physical state updates and attribute-specific
preservation requirements with regions of the relevant
entities and interaction effects.

\begin{figure*}[!t]
    \centering

    \IfFileExists{figs/model.pdf}{%
        \includegraphics[width=1.0\textwidth]{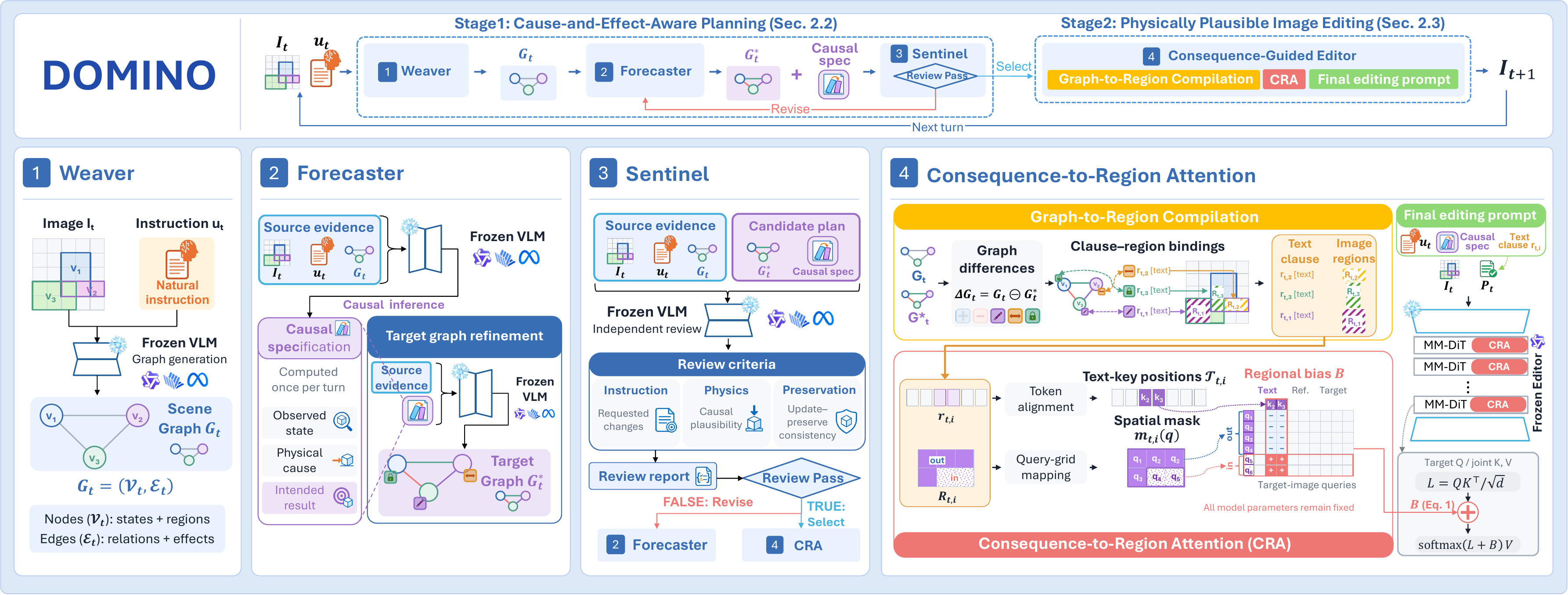}%
    }{%
        \begingroup
        \setlength{\fboxsep}{0pt}
        \fcolorbox{black!30}{black!5}{%
            \begin{minipage}[c][8cm][c]
                {\dimexpr\textwidth-2\fboxrule\relax}
                \centering
                {\large\bfseries Model Overview Placeholder}
                \par\medskip
                {\small Place the final figure in
                \texttt{figs/model.pdf}.}
            \end{minipage}%
        }
        \endgroup
    }

    \caption{
        \textbf{Overview of DOMINO for physically plausible
        multi-turn image editing at turn $t$.}
        (1) The Weaver extracts a scene graph representing
        entities, physical states, and interactions.
        (2) The Forecaster infers physical consequences and
        constructs an editing plan specifying physical state
        updates and preservation requirements.
        (3) The Sentinel reviews the plan and provides
        feedback for refinement.
        (4) The Consequence-Guided Editor uses
        Consequence-to-Region Attention (CRA) to guide a frozen
        image editor with text--region associations.
        The output $I_{t+1}$ becomes the next turn's input.
        Appendix~\ref{sec:supp_method_details}.
    }
    \label{fig:model}
\end{figure*}

To address these planning and spatial execution requirements,
we propose \modelname{} as a
\textbf{cause-and-effect-aware graph-based agent}
for multi-turn image editing (Fig.~\ref{fig:model}).
The method operates in two stages: planning by the agent
and image editing with a frozen editor.
In the Stage~1 planning phase, the agent
represents the current scene as a graph,
uses cause-and-effect reasoning to infer how the requested
change to one entity affects the physical states of related
entities, and constructs
an editing plan for review and refinement
(Sec.~\ref{sec:planning}).
In the Stage~2 image editing phase, we propose
\textbf{Consequence-to-Region Attention (CRA)}
as an attention-control module within the
Consequence-Guided Editor.
CRA converts the physical state updates and attribute-specific
preservation requirements inferred during planning into
spatial attention guidance.
CRA associates these requirements with regions of the relevant
entities and interaction effects and uses the resulting
consequence-to-region associations to guide a frozen image
editor~\cite{teamqwen}
(Sec.~\ref{sec:attention_control}).
These explicit associations enable \textbf{interactive control},
allowing users to inspect the editing plan and adjust the
image-region boxes used for spatial guidance.

\subsection{Stage~1: Cause-and-Effect-Aware Planning}
\label{sec:planning}

At editing turn $t$, \modelname{} plans an edit from the
current image $I_t$ and a natural instruction $u_t$ through
three components (Fig.~\ref{fig:model}).
All three use the same frozen vision-language model
(VLM)~\cite{yang2025qwen3}.
We provide the detailed algorithm and prompts in Appendix~\ref{sec:supp_algorithm_prompts}.

\paragraph{Weaver: scene graph construction.}

The Weaver extracts a scene graph
$G_t=(\mathcal{V}_t,\mathcal{E}_t)$ from $I_t$ and $u_t$.
Here, $\mathcal{V}_t$ and $\mathcal{E}_t$ denote the node
and edge sets, respectively.
\begin{itemize}
    \setlength{\itemsep}{0pt}
    \setlength{\parskip}{0pt}
    \item
    \textbf{Entity nodes.}
    Each node $v \in \mathcal{V}_t$ represents a scene
    entity and records its ID, with appearance, material,
    and physical states stored as attributes.
    
    \item
    \textbf{Interaction edges.}
    Each edge $e \in \mathcal{E}_t$ encodes a physical
    interaction between two entities and records their
    IDs, the interaction type, and effect attributes.
    
    \item
    \textbf{Spatial fields.}
    These fields associate entities and interaction effects
    with image regions, providing the spatial information
    used to apply the editing plan.
    
\end{itemize}
Detailed field definitions and conventions are provided
in Appendix~\ref{sec:supp_scene_graph}.

\paragraph{Forecaster: causal editing plan.}

The Forecaster infers the physical consequences of $u_t$
from $I_t$ and $G_t$.
A causal specification records the relevant entities and
attributes, together with the initial state, physical cause,
and resulting state of each physical consequence.
The Forecaster then constructs an editing plan that represents
the intended scene with a target graph $G_t^*$ and specifies physical
state updates and preservation requirements.
Entities shared by the current and target graphs retain their
IDs, and each physical consequence is linked to the target-graph entries
that realize it.
Before review, we convert the graph differences and spatially
grounded preservation requirements into text clauses paired
with image regions.
These pairs form the regional editing conditions used for
plan review and subsequent image editing
(Sec.~\ref{sec:attention_control}).

\paragraph{Sentinel: plan review.}

The Sentinel reviews the plan through a separate
call to the same VLM.
Using $I_t$, $u_t$, and $G_t$, it checks the target graph $G_t^*$ and
regional editing conditions for instruction following,
physical plausibility, and consistency with the causal
specification and preservation requirements.
The Forecaster uses this feedback to refine the plan
before image editing.
Detailed verification criteria and prompts are provided
in Appendix~\ref{sec:supp_plan_review}.

\begin{figure*}[!t]
    \centering
    \includegraphics[width=1.0\textwidth]{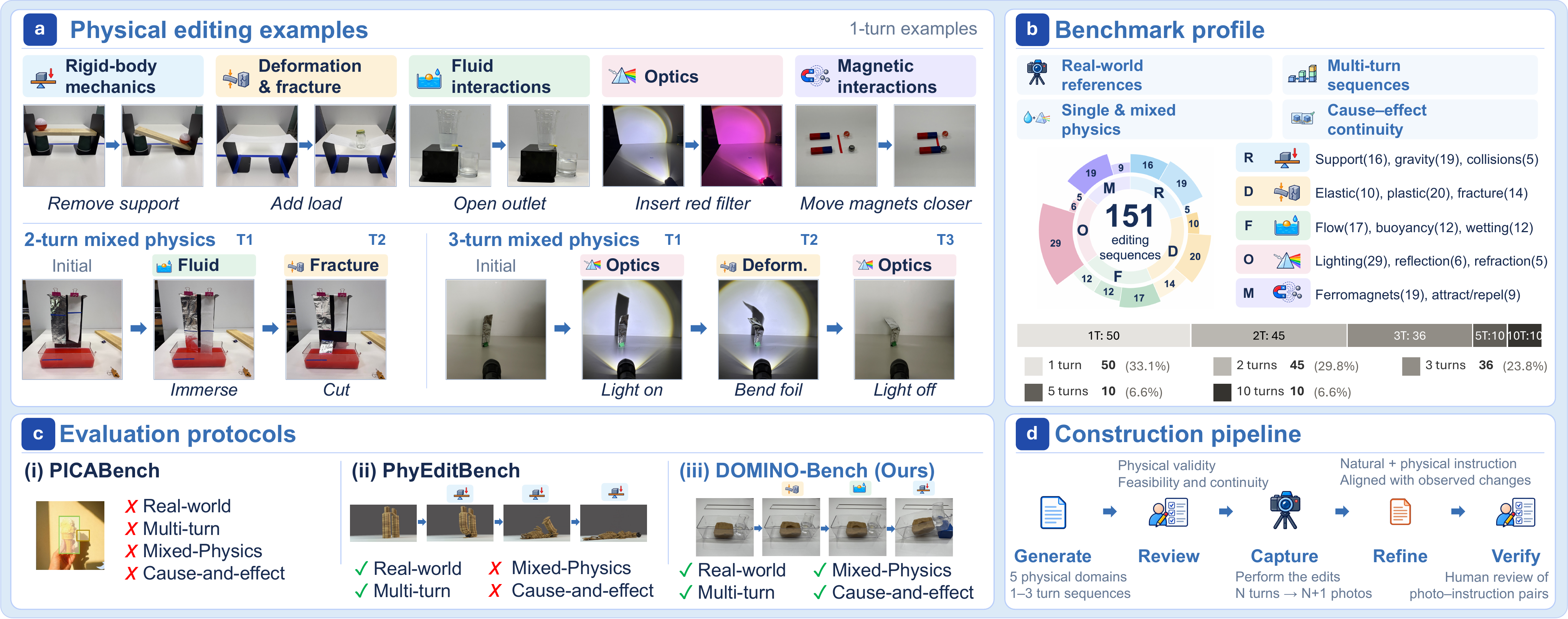}
    \caption{
        \textbf{Overview of DOMINO-Bench.}
        Real-world photographs show five physical domains,
        one-turn edits, and two- and three-turn mixed-physics sequences,
        emphasizing how earlier physical changes persist across turns.
        The radial chart summarizes reviewed scenario candidates:
        angular widths encode proportions of category assignments,
        and radial heights encode subcategory counts.
        The bottom panel compares benchmark examples and evaluation
        protocols, highlighting our use of each generated output
        as the next input.
        The rightmost column summarizes the dataset construction pipeline.
        Additional examples from \benchmarkname{} are provided in Appendix~\ref{sec:supp_benchmark_examples}.
    }
    \label{fig:benchmark_main}
\end{figure*}


\subsection{Stage~2: Physically Plausible Image Editing}
\label{sec:attention_control}

At editing turn $t$, the Consequence-Guided Editor applies
the final editing plan to $I_t$ using a frozen image
editor~\cite{teamqwen}, producing $I_{t+1}$.
We propose \textbf{Consequence-to-Region Attention (CRA)}
as an attention-control module that converts the plan's
regional conditions into spatial attention guidance.

The Consequence-Guided Editor receives the current graph
$G_t$ and the reviewed editing plan from Stage~1,
including the target graph $G_t^*$ and preservation requirements.
Let $\Delta G_t = G_t \ominus G_t^*$ denote the changes
from the current graph $G_t$ to the target graph $G_t^*$,
where $\ominus$ identifies changes to entities, attributes,
and interactions.
Before plan review (Sec.~\ref{sec:planning}),
Graph-to-Region Compilation converts these changes
and preservation requirements into text clauses $r_{t,i}$
paired with image regions $R_{t,i}$,
where $i$ indexes the clauses.
Each pair associates a physical state update or preservation
requirement with the region of the relevant entity or
interaction effect.
The text clauses are included in the final prompt $P_t$
supplied to the frozen image editor.
We detail region selection for each editing operation
and final prompt construction in Appendix~\ref{sec:supp_graph_to_region}.

CRA encourages attention to each text clause within its
associated image region and discourages attention to that
clause outside the region.
We omit the editing-turn index $t$ below for clarity.
For one attention head, let $\boldsymbol{Q}$ denote target-image
queries and $\boldsymbol{K},\boldsymbol{V}$ denote joint text,
reference-image, and target-image keys and values.
The original attention logits are
$\boldsymbol{L}=\boldsymbol{Q}\boldsymbol{K}^{\top}/\sqrt{d}$,
where $d$ is the head dimension.
For clause $r_i$, let $\mathcal{T}_i$ denote its text-key
positions and $c_i\in[0,1]$ its confidence weight.
Let $m_i(q)=1$ when query $q$ lies within $R_i$, and $0$
otherwise.
CRA adds the regional bias:
{
\begin{equation}
\begin{aligned}
B_{qk}
&= \sum_i \alpha c_i\,1[k\in\mathcal{T}_i] \\
&\quad\cdot\left[\beta_{\mathrm{in}}m_i(q)
-\beta_{\mathrm{out}}\bigl(1-m_i(q)\bigr)\right].
\end{aligned}
\label{eq:cra_bias}
\end{equation}
}
In Eq.~\ref{eq:cra_bias}, $\alpha$ controls the overall
strength, while $\beta_{\mathrm{in}},\beta_{\mathrm{out}}>0$
control the inside and outside biases.
Update and preservation clauses use the same mechanism
and can share an image region.

We correct the native attention output using the difference
between attention outputs computed with and without the
regional bias.
Both computations span all joint keys and retain the original
padding mask.
The correction applies only to target-image queries;
text and reference-image queries retain their native attention.
The full formulation is provided in Appendix~\ref{sec:supp_graph_to_region}.

Users can inspect the editing plan and adjust the associated
image-region boxes to control where its conditions apply (Sec.~\ref{sec:control}).
Moving a box changes $R_i$ and the query locations selected
by $m_i(q)$ in Eq.~\ref{eq:cra_bias}, directing the same
text clause to the revised region.

\section{\benchmarkname{}}
\label{sec:benchmark}

\definecolor{bestbg}{HTML}{E2F0D9}
\definecolor{secondbg}{HTML}{E3E8ED}

\newcommand{\best}[1]{\cellcolor{bestbg}{\textbf{#1}}}
\newcommand{\second}[1]{\cellcolor{secondbg}{\underline{#1}}}

\begin{table*}[!t]
\centering
\caption{
    \textbf{Quantitative comparison on \benchmarkname{},
    MagicBrush, and PICABench.}
    For each method and metric, we average valid scores over
    successfully generated outputs.
    MB: MultiBanana Physical; KRIS: KRIS Knowledge;
    ER: EditReward.
    KRIS Knowledge is evaluated with benchmark explanations on
    \benchmarkname{}, MagicBrush~\cite{zhang2023magicbrush}
    and PICABench~\cite{pu2026picabench}.
    \underline{All metrics are higher-is-better.}
    The best and second-best scores in each column are shown in
    \textcolor{bestbg}{\rule{0.8em}{0.8em}}~\textbf{light green}
    and
    \textcolor{secondbg}{\rule{0.8em}{0.8em}}~\underline{light gray},
    respectively.
}
\label{tab:main_comparison}
\footnotesize
\setlength{\tabcolsep}{3pt}
\renewcommand{\arraystretch}{1.15}
\resizebox{\linewidth}{!}{%
\begin{tabular}{>{\centering\arraybackslash}m{2.65cm} l c *{11}{c}}
\toprule
\multirow{2}{*}{\textbf{Category}}
& \multirow{2}{*}{\textbf{Method}}
& \multirow{2}{*}{\shortstack{\textbf{Training}\\\textbf{free}}}
& \multicolumn{5}{c}{\textbf{\benchmarkname{} (Ours)}}
& \multicolumn{3}{c}{\textbf{MagicBrush (NeurIPS'23)}}
& \multicolumn{3}{c}{\textbf{PICABench (ICLR'26)}}  \\
\cmidrule(lr){4-8}
\cmidrule(lr){9-11}
\cmidrule(l){12-14} &  &  & MB & KRIS & ER & CLIP-I & CLIP-T & MB & KRIS & ER & MB & KRIS & ER \\
\midrule
\multirow{2}{=}{\centering Image\\editing} & DreamOmni2~\cite{Xia_2026_CVPR} (CVPR'26) & \xmark & 8.9073 & 2.6424 & 0.3671 & \second{0.9506} & 0.2327 & 8.9934 & 3.5033 & 0.6983 & 8.3046 & 3.6093 & 0.4736 \\
 & CoCoEdit~\cite{wu2026cocoedit} (ICML'26) & \xmark & 7.9934 & 3.5894 & 0.3887 & 0.8950 & 0.2339 & 9.2517 & \best{4.5099} & \second{1.3655} & 9.0530 & \second{4.5762} & \best{1.1821} \\
\midrule
\multirow{2}{=}{\centering Multi-turn\\editing} & FreqEdit~\cite{liao2026freqedit} (CVPR'26) & \cmark & \second{9.4305} & 3.1325 & 0.4539 & \best{0.9532} & 0.2303 & 9.1854 & 3.3907 & 0.8203 & 8.8146 & 3.5828 & 0.5285 \\
 & MTC~\cite{zhou2025multi} (ICCV'25) & \cmark & 5.1854 & 2.0795 & $-0.2404$ & 0.8299 & \best{0.2452} & 5.5828 & 2.0927 & $-0.7410$ & 4.3775 & 1.6159 & $-0.8592$ \\
\midrule
\multirow{2}{*}{\shortstack{Physically plausible\\editing}} & ChronoEdit~\cite{wu2026chronoedit} (ICLR'26) & \xmark & 7.5497 & 2.4106 & 0.1001 & 0.9195 & 0.2354 & 7.7682 & 2.6424 & 0.0596 & 7.8940 & 3.0728 & 0.0710 \\
 & PhysicEdit~\cite{zhao2026from} (ICML'26) & \xmark & 7.9603 & 3.1788 & 0.1738 & 0.9047 & 0.2328 & 8.3510 & 4.0662 & 0.7581 & 9.0728 & 4.5364 & 0.9950 \\
\midrule
\multirow{2}{=}{\centering Graph-based\\editing} & SGEdit~\cite{SGEdit} (SIGGRAPH Asia'24) & \xmark & 5.4967 & 2.9470 & $-0.3917$ & 0.8725 & 0.2144 & 6.9404 & 2.8411 & $-0.0788$ & 5.0861 & 3.3709 & $-0.3587$ \\
 & VENUS~\cite{VENUS} (MMM'26) & \cmark & 7.6159 & 2.5894 & $-0.0467$ & 0.9249 & 0.2335 & 8.2649 & 2.1523 & $-0.2649$ & 7.9868 & 2.1391 & $-0.6153$ \\
\midrule
\multirow{2}{=}{\centering Agent-based\\editing} & ReasonEdit-S~\cite{yin2026reasonedit} (CVPR'26) & \xmark & 9.1523 & \second{4.1457} & 0.3273 & 0.9082 & 0.2291 & \second{9.3576} & 4.4967 & 1.3534 & 9.0861 & 4.1854 & 0.9525 \\
 & RePlan~\cite{qu2026replan} (ECCV'26) & \xmark & 9.1060 & 3.2119 & \second{0.4930} & 0.9504 & 0.2273 & 8.6821 & 3.8874 & 0.9159 & \second{9.2781} & 4.2053 & 1.0071 \\
\midrule
\textbf{Cause-and-effect editing} & \textbf{\modelname{} (Ours)} & \cmark & \best{9.6093} & \best{4.5629} & \best{0.5501} & 0.9320 & \second{0.2371} & \best{9.3576} & \second{4.5033} & \best{1.3753} & \best{9.4305} & \best{4.6755} & \second{1.1323} \\

\bottomrule
\end{tabular}%
}
\end{table*}

\subsection{Motivation}
\label{subsec:bm_motiv}

In physically plausible multi-turn editing, earlier state changes can affect
subsequent physical outcomes.
Evaluation must therefore assess cause-and-effect continuity
across multi-turns.
Mixed-physics scenarios further test whether models account
for interactions between effects from different physical domains.
PICABench~\cite{pu2026picabench} evaluates single-turn physical
plausibility without captured ground-truth outcomes.
PhyEditBench~\cite{guo2026phyeditbench} samples successive states
from real-world videos to evaluate the progression
of a physical process.
However, this setting does not explicitly target interactions
between physical effects introduced by distinct editing
interventions or cause-and-effect dependencies between
those interventions (Fig.~\ref{fig:benchmark_main}(c)).
To address this evaluation gap, we construct \benchmarkname{}
to evaluate physical consequences and state consistency across
successive edits using photographed real-world outcomes
as references.

\subsection{Benchmark Characteristics}
\label{sec:benchmark_characteristics}

As summarized in Fig.~\ref{fig:benchmark_main}(b),
\benchmarkname{} provides \textbf{real-world references}
of physical outcomes and \textbf{multi-turn sequences}
spanning short and long editing histories.
\textbf{Single- and mixed-physics scenarios} cover individual
effects and interactions between effects.
These sequences test \textbf{cause-and-effect continuity}
by assessing how earlier changes affect later edits
and which states should persist.
The detailed construction pipeline is provided in Appendix~\ref{sec:supp_benchmark_construction}.

\section{Experiments}
\label{sec:exp}

We use frozen Qwen3-VL-8B~\cite{yang2025qwen3}
for all three Stage~1 subagents and a frozen Qwen image
editor~\cite{teamqwen} for Stage~2.
We evaluate \modelname{} on \benchmarkname{},
MagicBrush~\cite{zhang2023magicbrush},
and PICABench~\cite{pu2026picabench}.
EditReward~\cite{wu2026editreward} (ER) assesses instruction
following and visual quality using a reward model trained
on human preferences.
We adopt MultiBanana's Physical Realism
score~\cite{oshima2026multibanana} (MB Physical) to assess
the physical plausibility of edited scenes.
KRIS-Bench's Knowledge Plausibility
score~\cite{wu2026kris} (KRIS Knowledge) assesses whether
edited images are consistent with real-world knowledge
and domain-specific principles.
CLIP-I and CLIP-T~\cite{pmlr-v139-radford21a} additionally
measure image similarity and image-text alignment,
respectively.
Implementation and evaluator settings are provided in
Appendix~\ref{sec:supp_experimental_details}.

\begin{figure*}[!t]
    \centering
    \includegraphics[width=\textwidth]{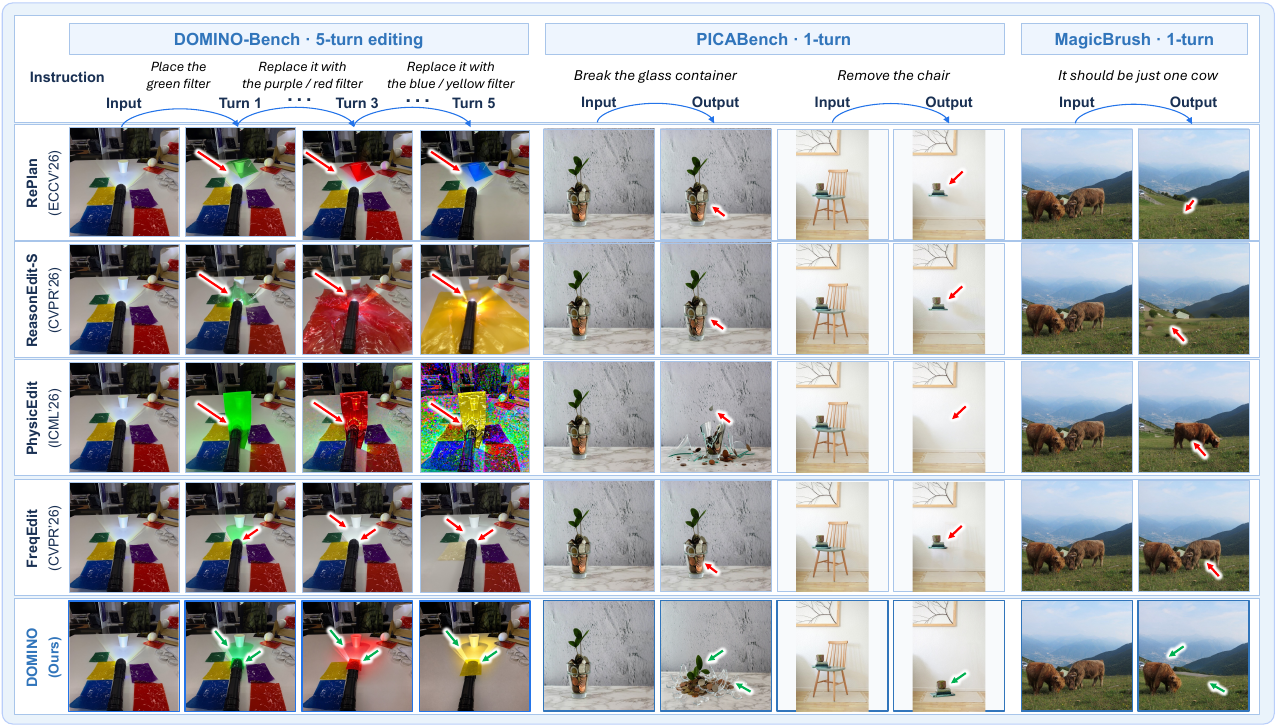}
    \caption{
        \textbf{Qualitative comparison across benchmarks
        and editing turns.}
        We compare results on \benchmarkname{}, MagicBrush,
        and PICABench.
        The left sequence shows selected turns from five
        successive color-filter edits; the remaining pairs
        show input and edited images.
        \modelname{} updates illumination across turns and
        preserves the supported objects while placing them
        on the floor after chair removal.
    }
    \label{fig:qualitative_comparison}

\end{figure*}

\subsection{Comparison with Existing Methods}
\label{sec:quan}

Across \benchmarkname{}, MagicBrush, and PICABench,
\modelname{} ranks among the top two in MB Physical,
KRIS Knowledge, and ER, leading all three metrics on
\benchmarkname{} (Tab.~\ref{tab:main_comparison}).
It exceeds SGEdit, VENUS, ReasonEdit-S, and RePlan on
these three metrics across all three benchmarks.
CoCoEdit leads KRIS Knowledge on MagicBrush and ER
on PICABench.
FreqEdit and MTC instead lead CLIP-I and CLIP-T on
\benchmarkname{}, respectively, showing that similarity
and alignment rankings differ from those of the editing
and physical evaluation metrics.

PhysicEdit ranks more competitively on single-turn PICABench
than on \benchmarkname{}.
PhysicEdit models physical state transitions using
video-derived priors~\cite{zhao2026from}, while ChronoEdit
models the source and edited images as the first and last
frames of a video~\cite{wu2026chronoedit}.
Our multi-turn evaluation additionally requires reasoning
from the physical states established by earlier edits
to determine which states to update and which to preserve
under each new intervention.

The chair-removal example in
Fig.~\ref{fig:qualitative_comparison} shows a
cause-and-effect relationship across interacting entities:
removing the chair also requires updating the positions
of the objects it supports.
\modelname{} preserves the cup and books and places them
on the floor.
ReasonEdit-S and RePlan leave these objects floating,
while PhysicEdit removes them.
In the MagicBrush example, \modelname{} removes one cow
while preserving the other and the surrounding scene.
FreqEdit retains both cows, whereas ReasonEdit-S and
RePlan remove both.

In the five-turn filter sequence
(Fig.~\ref{fig:qualitative_comparison}), each filter
replacement requires updating the resulting illumination.
\modelname{} updates illumination with each filter color,
whereas FreqEdit returns to white illumination and
PhysicEdit develops severe visual artifacts in later turns.

We collect separate five-point human ratings for instruction
following and physical plausibility in 41 evaluation sessions
on 30 cases, with 10 cases per benchmark
(Tab.~\ref{tab:user_study}).
\modelname{} receives the highest mean ratings on both
criteria overall and on every benchmark.
Its margin over the strongest baseline is largest on
\benchmarkname{} for both criteria.
ReasonEdit-S ranks second overall in instruction following,
whereas RePlan ranks second in physical plausibility,
showing that the two criteria yield different rankings.
Together, the automatic metrics, qualitative examples,
and human ratings support \modelname{}'s ability to follow
editing instructions while updating physical states and
preserving relevant scene content on the evaluated cases.
The study interface is provided in
Appendix~\ref{sec:supp_human_evaluation}.
Additional qualitative analysis and a comparison of automatic
evaluation results with human ratings are provided in
Appendix~\ref{sec:supp_qualitative_results} and
Appendix~\ref{sec:supp_auto_human_comparison}, respectively.

\begin{table}[!htbp]
    \centering
    \caption{
        \textbf{Human evaluation across benchmarks.}
        Mean human ratings (1--5) from 41 evaluation sessions
        on 30 cases, with 10 cases per benchmark.
        IF: instruction following; PP: physical plausibility.
        Higher is better.
        The best and second-best scores in each row are shown in
        \textcolor{bestbg}{\rule{0.8em}{0.8em}}~\textbf{light green}
        and
        \textcolor{secondbg}{\rule{0.8em}{0.8em}}~\underline{light gray}.
    }
    \label{tab:user_study}
    \footnotesize
    \setlength{\tabcolsep}{2.5pt}
    \renewcommand{\arraystretch}{1.12}
    \resizebox{\columnwidth}{!}{%
    \begin{tabular}{@{}llcccc@{}}
        \toprule
        \textbf{Benchmark}
        &
        & \shortstack{\textbf{ReasonEdit-S}\\
                      \scriptsize (CVPR'26)}
        & \shortstack{\textbf{FreqEdit}\\
                      \scriptsize (CVPR'26)}
        & \shortstack{\textbf{RePlan}\\
                      \scriptsize (ECCV'26)}
        & \shortstack{\textbf{\modelname{}}\\
                      \scriptsize (Ours)} \\
        \midrule

        \multirow{2}{*}{\textbf{Overall}}
        & IF
        & \second{2.7805} & 2.1951 & 2.3659 & \best{4.4146} \\
        & PP
        & 3.2439 & 3.0244 & \second{3.2927} & \best{4.3415} \\
        \midrule

        \multirow{2}{*}{\benchmarkname{}}
        & IF
        & \second{2.2195} & 1.8049 & 1.7317 & \best{4.6341} \\
        & PP
        & 2.9024 & 3.0732 & \second{3.0732} & \best{4.5366} \\
        \midrule

        \multirow{2}{*}{MagicBrush}
        & IF
        & 2.9512 & 2.7805 & \second{2.9512} & \best{4.4146} \\
        & PP
        & 3.3659 & 3.3415 & \second{3.3902} & \best{4.2683} \\
        \midrule

        \multirow{2}{*}{PICABench}
        & IF
        & \second{3.1463} & 2.0488 & 2.4146 & \best{4.1951} \\
        & PP
        & \second{3.4878} & 2.6829 & 3.3902 & \best{4.2439} \\
        \bottomrule
    \end{tabular}%
    }
\end{table}


\subsection{Ablation Studies}
\label{sec:ab_study}

We conduct three ablation studies.
\textit{(i) Agent planning and CRA.}
We compare the editor-only baseline, Stage~1 without CRA,
and the full method, keeping the editing plan and final
prompt identical between the latter two.
\textit{(ii) Plan review versus image review.}
We compare no review, plan review before editing,
and image review after editing, retaining the Weaver,
Forecaster, and CRA.
\textit{(iii) Agent models.}
We replace the VLM shared by all three Stage~1 subagents
while keeping the Consequence-Guided Editor, including CRA,
and the remaining pipeline fixed.
\paragraph{Effect of Agent Planning and CRA.}
Adding Stage~1 improves MB Physical, KRIS Knowledge, and ER
(Tab.~\ref{tab:ab_framework}),
indicating that planning and review provide useful conditions
for the frozen image editor.
Enabling CRA further improves MB Physical and KRIS Knowledge
but decreases ER.
With the plan and final prompt unchanged, these gains support
the benefit of spatial attention guidance when applying
physical state updates and preservation requirements,
although the gains do not extend to every measure
of editing quality.

\begin{table}[!htbp]
    \centering
    \caption{
        \textbf{Ablation of Stage~1 planning (Agent) and CRA.}
        Higher is better for all metrics.
        The best and second-best scores are shown in
        \textcolor{bestbg}{\rule{0.8em}{0.8em}}~\textbf{light green}
        and
        \textcolor{secondbg}{\rule{0.8em}{0.8em}}~\underline{light gray}.
    }
    \label{tab:ab_framework}

    \small
    \setlength{\tabcolsep}{4pt}
    \renewcommand{\arraystretch}{1.1}

    {%
        \begin{tabularx}{\columnwidth}{@{}Xccrrr@{}}
            \toprule
            \textbf{Variant} & \textbf{Agent} & \textbf{CRA}
            & \textbf{MB} $\uparrow$
            & \textbf{KRIS} $\uparrow$
            & \textbf{ER} $\uparrow$ \\
            \midrule

            Baseline & \xmark & \xmark
            & 9.4783 & 3.9674 & \second{0.5634} \\

            w/o CRA & \cmark & \xmark
            & \second{9.5154} & \second{4.4030}
            & \best{0.5946} \\

            \textbf{Full} & \cmark & \cmark
            & \best{9.6093} & \best{4.5629} & 0.5501 \\

            \bottomrule
        \end{tabularx}%
    }
\end{table}

\paragraph{Plan Review versus Image Review.}
Plan review improves MB Physical and KRIS Knowledge over
no review at similar runtime, although ER decreases
(Tab.~\ref{tab:ab_sentinel}).
Image review achieves slightly higher MB Physical and ER
than plan review but lower KRIS Knowledge, with substantially
more runtime (Sec.~\ref{sec:com_cost}).
In image review, correcting errors identified by the Sentinel
still depends on the frozen image editor, so the Sentinel's
feedback alone does not guarantee successful correction.
The mixed gains and higher runtime support using the Sentinel
to review the editing plan before execution as a more efficient
choice in this comparison, achieving the highest KRIS Knowledge
and nearly the same MB Physical as image review.

\begin{table}[!htbp]
    \centering
    \caption{
        \textbf{Sentinel ablation.}
        Weaver, Forecaster, and CRA are retained in all variants.
        Time reports average runtime in seconds, with percentages
        normalized to Full (100\%) in parentheses.
        The best and second-best values are shown in
        \textcolor{bestbg}{\rule{0.8em}{0.8em}}~\textbf{light green}
        and
        \textcolor{secondbg}{\rule{0.8em}{0.8em}}~\underline{light gray}.
    }
    \label{tab:ab_sentinel}

    \footnotesize
    \setlength{\tabcolsep}{2.5pt}
    \renewcommand{\arraystretch}{1.1}

    \resizebox{\columnwidth}{!}{%
        \begin{tabular}{@{}llrrrr@{}}
            \toprule
            \textbf{Variant}
            & \shortstack{\textbf{Review}\\\textbf{target}}
            & \textbf{MB} $\uparrow$
            & \textbf{KRIS} $\uparrow$
            & \textbf{ER} $\uparrow$
            & \textbf{Time (s)} $\downarrow$ \\
            \midrule

            w/o Sentinel & None
            & 9.3832 & \second{4.4664} & \best{0.5750}
            & \best{126.7 (98.3\%)} \\

            Image Review & Edited image
            & \best{9.6176} & 4.3824 & \second{0.5581}
            & 232.7 (180.5\%) \\

            \textbf{Full} & Editing plan
            & \second{9.6093} & \best{4.5629} & 0.5501
            & \second{128.9 (100.0\%)} \\

            \bottomrule
        \end{tabular}%
    }
\end{table}

\paragraph{Choice of Agent Models.}
In Stage~1, the Weaver constructs a scene graph, and the
Forecaster uses the graph to infer physical consequences
and construct an editing plan.
The Sentinel reviews the plan's physical state updates and
attribute-specific preservation requirements before image editing.
The smaller local VLMs retain much of the default (8B) model's
performance (Tab.~\ref{tab:ab_models}), with the 4B variant
achieving particularly close ER and KRIS Knowledge scores.
GPT-6-Astra further improves all three metrics.
These results support using compact local VLMs to construct
and review cause-and-effect-aware editing plans, while
showing that replacing the Stage~1 model can further improve
editing quality and physical plausibility.

\begin{table}[!htbp]
    \centering
    \caption{
        \textbf{Stage~1 VLM ablation on \benchmarkname{}.}
        All three Stage~1 subagents use the replacement VLM,
        while the Stage~2 image editor remains fixed.
        The best and second-best scores across all variants
        are shown in
        \textcolor{bestbg}{\rule{0.8em}{0.8em}}~\textbf{light green}
        and
        \textcolor{secondbg}{\rule{0.8em}{0.8em}}~\underline{light gray}.
    }
    \label{tab:ab_models}

    \small
    \setlength{\tabcolsep}{4pt}
    \renewcommand{\arraystretch}{1.1}

    {%
        \begin{tabularx}{\columnwidth}{@{}Xrrr@{}}
            \toprule
            \textbf{Stage~1 VLM}
            & \textbf{MB} $\uparrow$
            & \textbf{KRIS} $\uparrow$
            & \textbf{ER} $\uparrow$ \\
            \midrule

            Qwen3-VL-2B-Instruct
            & 9.1889 & 4.2889 & 0.4969 \\

            Qwen3-VL-4B-Instruct
            & 9.3677 & 4.5130 & 0.5428 \\

            Qwen3-VL-8B-Instruct (default)
            & \second{9.6093} & \second{4.5629}
            & \second{0.5501} \\
            \midrule

            GPT-6-Luna
            & 9.1429 & 4.5238 & 0.5110 \\

            GPT-6-Astra
            & \best{9.7677} & \best{4.7271} & \best{0.5961} \\

            \bottomrule
        \end{tabularx}%
    }
\end{table}
\subsection{Computational Cost}
\label{sec:com_cost}
As shown in Fig.~\ref{fig:teaser}(b), \modelname{} achieves
the highest KRIS Knowledge and ER scores among the compared
methods while maintaining competitive runtime.
This balance supports the practicality of cause-and-effect-aware
planning for multi-turn image editing.

\subsection{Interactive Control}
\label{sec:control}
\begin{figure}[!htbp]
    \centering
    \includegraphics[width=\linewidth]{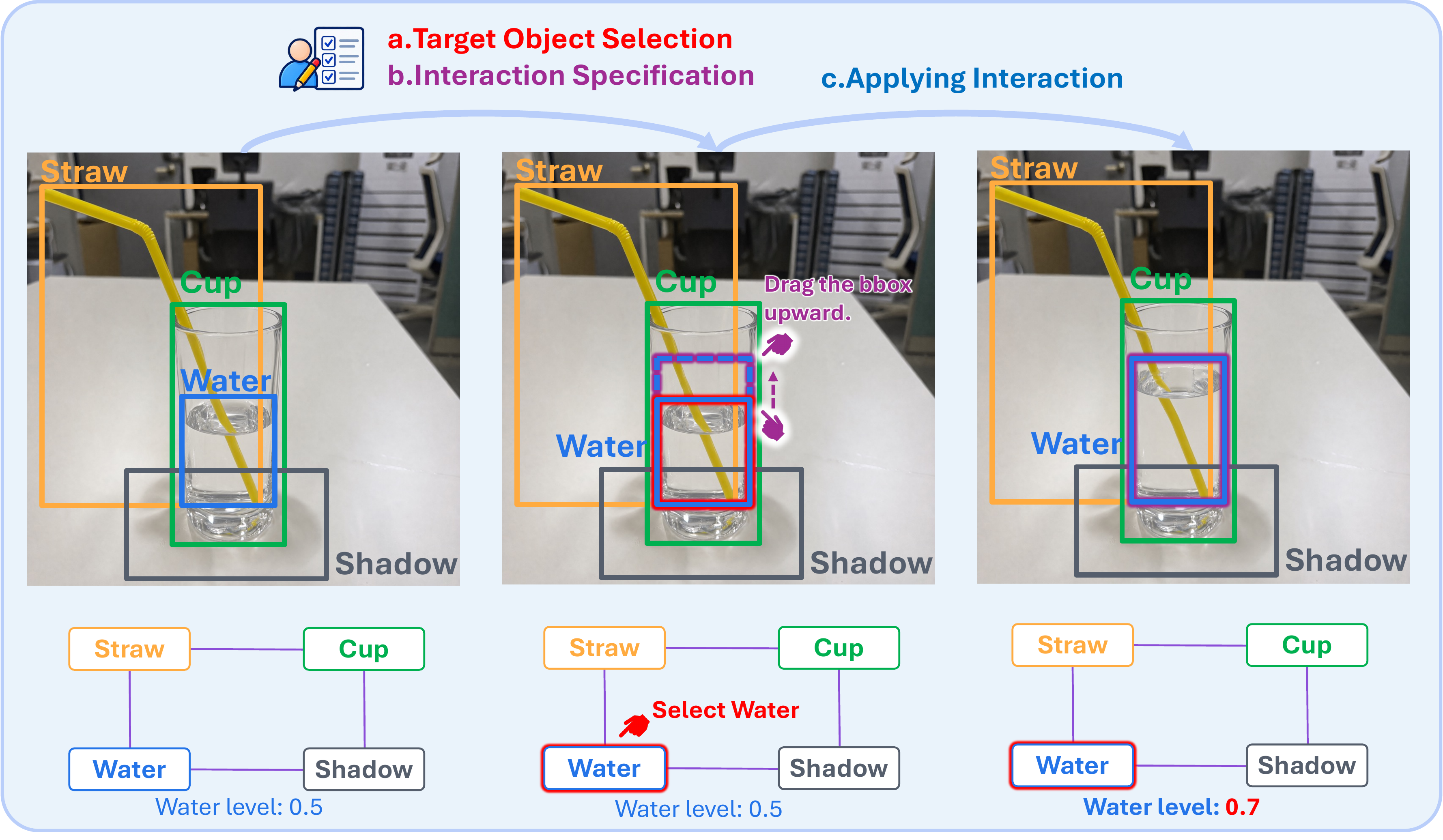}
    \caption{
        \textbf{Interactive control through CRA.}
        The user selects the water entity and moves its
        image-region box upward.
        CRA uses the revised region to guide the frozen     
        editor to raise the water level.
    }
    \label{fig:interactive_multiturn}
\end{figure}

Our CRA enables \textbf{interactive control} through the
consequence-to-region associations in the editing plan.
Users can inspect the plan, select an entity in the scene
graph, and adjust its associated image-region box to specify
where a planned physical state update should apply.
In Fig.~\ref{fig:interactive_multiturn}, the user selects
the water entity and moves its box upward to request
a higher water level.
CRA uses the revised region to guide the frozen image editor
through spatial attention control.
The scene graph therefore supports both cause-and-effect
reasoning and user control over where the resulting
physical state updates apply.
Additional examples of interactive control are provided
in Appendix~\ref{sec:supp_interactive_control}.

\section{Conclusion}
\label{sec:conclusion}

We have proposed \modelname{}, a training-free,
cause-and-effect-aware graph-based agent for physically
plausible multi-turn image editing.
The agent constructs and reviews editing plans, while CRA
associates their physical state updates and attribute-specific
preservation requirements with image regions to guide
a frozen editor and enable interactive control.
We have also constructed \benchmarkname{} from photographed
outcomes of sequential physical interventions to evaluate
cause-and-effect continuity.
Experiments have demonstrated competitive editing quality
and physical plausibility.
\RestoreCVPRSectionFont


\clearpage
\pdfbookmark[1]{References}{domino-references}
{\small
\bibliographystyle{ieeenat_fullname}
\bibliography{main}
}

\clearpage
\appendix
\providecommand{\suppfigplaceholder}[2]{}

\renewcommand{\suppfigplaceholder}[2]{%
    \begingroup
    \setlength{\fboxrule}{0.8pt}%
    \setlength{\fboxsep}{0pt}%
    \fcolorbox{black}{white}{%
        \parbox[c][#1][c]{%
            \dimexpr\linewidth-2\fboxsep-2\fboxrule\relax
        }{%
            \centering
            \color{black}%
            \normalsize
            #2\\[0.5em]
            Placeholder
        }%
    }%
    \endgroup
}

\clearpage
\onecolumn
\setcounter{page}{1}

\begin{center}
    {\Large\bfseries \PaperTitle\par}
    \vspace{0.5em}
\end{center}

\vspace{0.5em}
\section*{Supplementary Contents}
\par\smallskip
\begingroup
\large
\DominoMainSectionFace 
\begin{itemize}
    \setlength{\itemsep}{8pt}
    \setlength{\parsep}{0pt}

    \item \cref{sec:supp_related_work}\quad
    \hyperref[sec:supp_related_work]{Related Work}

    \item \cref{sec:supp_benchmark_details}\quad
    \hyperref[sec:supp_benchmark_details]{Benchmark Details}
    \begin{itemize}
        \setlength{\itemsep}{3pt}
        \setlength{\parsep}{0pt}

        \item \cref{sec:supp_benchmark_construction}\quad
        \hyperref[sec:supp_benchmark_construction]{Benchmark Construction Pipeline}

        \item \cref{sec:supp_benchmark_examples}\quad
        \hyperref[sec:supp_benchmark_examples]{Additional Examples from DOMINO-Bench}
    \end{itemize}

    \item \cref{sec:supp_method_details}\quad
    \hyperref[sec:supp_method_details]{Method Details}
    \begin{itemize}
        \setlength{\itemsep}{3pt}
        \setlength{\parsep}{0pt}

        \item \cref{sec:supp_algorithm_prompts}\quad
        \hyperref[sec:supp_algorithm_prompts]{Overall Algorithm and Agent Prompts}

        \item \cref{sec:supp_scene_graph}\quad
        \hyperref[sec:supp_scene_graph]{Scene Graph Representation}

        \item \cref{sec:supp_plan_review}\quad
        \hyperref[sec:supp_plan_review]{Plan Review and Refinement}

        \item \cref{sec:supp_graph_to_region}\quad
        \hyperref[sec:supp_graph_to_region]{Graph-to-Region Compilation and CRA}

        \item \cref{sec:supp_interactive_control}\quad
        \hyperref[sec:supp_interactive_control]{Interactive Control}
    \end{itemize}

    \item \cref{sec:supp_experimental_details}\quad
    \hyperref[sec:supp_experimental_details]{Experimental Settings and Evaluation Protocols}
    \begin{itemize}
        \setlength{\itemsep}{3pt}
        \setlength{\parsep}{0pt}

        \item \cref{sec:supp_implementation}\quad
        \hyperref[sec:supp_implementation]{Implementation and Baseline Settings}

        \item \cref{sec:supp_automatic_evaluation}\quad
        \hyperref[sec:supp_automatic_evaluation]{Automatic Evaluation}

        \item \cref{sec:supp_human_evaluation}\quad
        \hyperref[sec:supp_human_evaluation]{Human Evaluation Protocol and Interface}

        \item \cref{sec:supp_auto_human_comparison}\quad
        \hyperref[sec:supp_auto_human_comparison]{Comparison of Automatic and Human Evaluation}
    \end{itemize}

    \item \cref{sec:supp_qualitative_results}\quad
    \hyperref[sec:supp_qualitative_results]{Additional Qualitative Results}
    \begin{itemize}
        \setlength{\itemsep}{3pt}
        \setlength{\parsep}{0pt}

        \item \cref{sec:supp_entity_consequences}\quad
        \hyperref[sec:supp_entity_consequences]{Physical Consequences across Interacting Entities}

        \item \cref{sec:supp_turn_dependencies}\quad
        \hyperref[sec:supp_turn_dependencies]{Causal Dependencies across Editing Turns}

        \item \cref{sec:supp_project_page}\quad
        \hyperref[sec:supp_project_page]{Interactive Demonstrations and Additional Qualitative Examples}
    \end{itemize}

\end{itemize}
\endgroup


\twocolumn[{%
    \noindent
    \begin{minipage}{\textwidth}
        \centering
        \includegraphics[width=\linewidth]{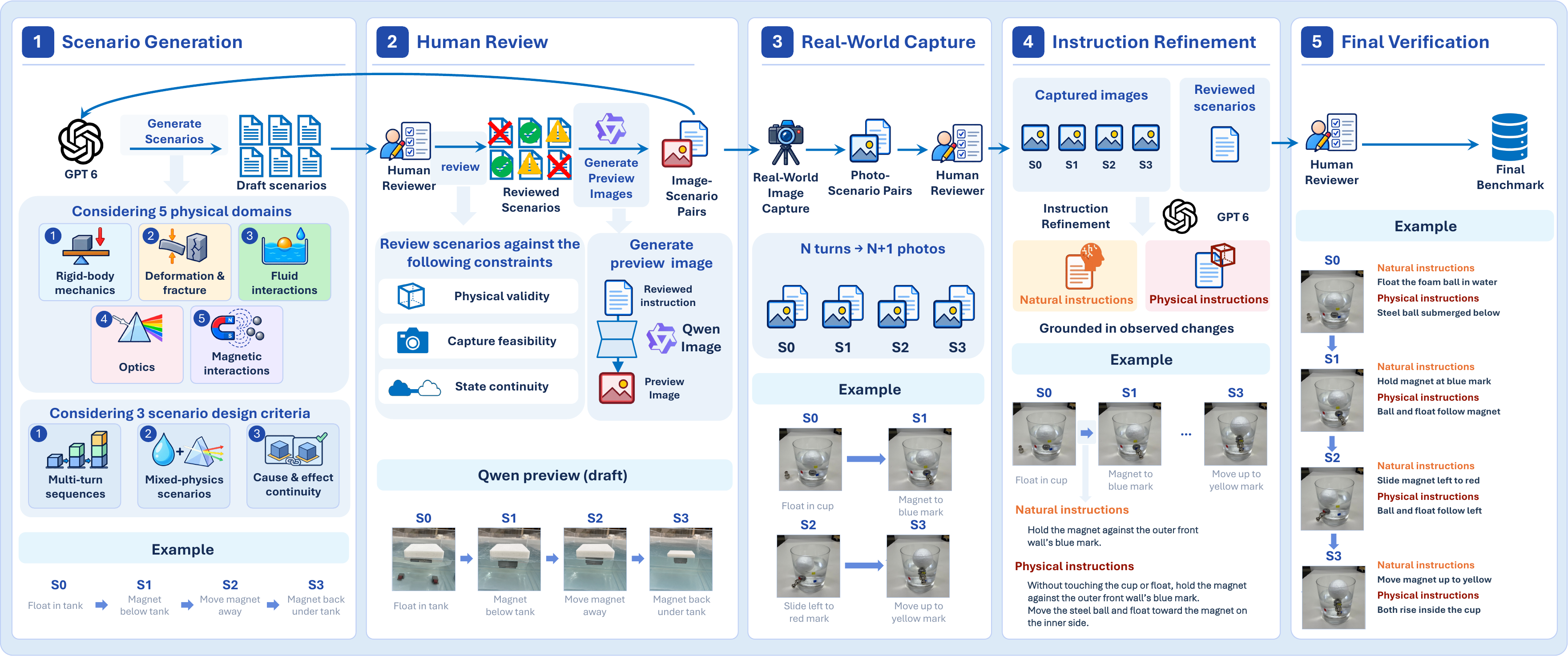}

        \captionsetup{
            hypcap=false,
            width=\linewidth
        }
        \captionof{figure}{
            \textbf{Construction of \benchmarkname{}.}
            \textbf{Stage~1:} GPT-6 drafts multi-turn scenarios across
            five physical domains.
            \textbf{Stage~2:} Human reviewers assess physical validity,
            capture feasibility, and state continuity, aided by
            Qwen-Image previews.
            \textbf{Stage~3:} Reviewed scenarios are executed with real
            objects and photographed, and scenario descriptions are
            revised to match the captured states.
            \textbf{Stage~4:} GPT-6 uses the captured photographs and
            revised descriptions to produce paired natural instructions
            for model input and physical instructions for evaluation.
            \textbf{Stage~5:} Human reviewers verify image--text alignment
            and state consistency before including each sequence in
            the benchmark.
            Each $N$-turn sequence contains $N+1$ photographs and
            $N$ natural--physical instruction pairs.
        }
        \label{fig:benchmark_pipeline}
    \end{minipage}
    \par
    \vspace{1em}
}]

\begin{figure*}[t]
    \centering
    \includegraphics[width=\textwidth]{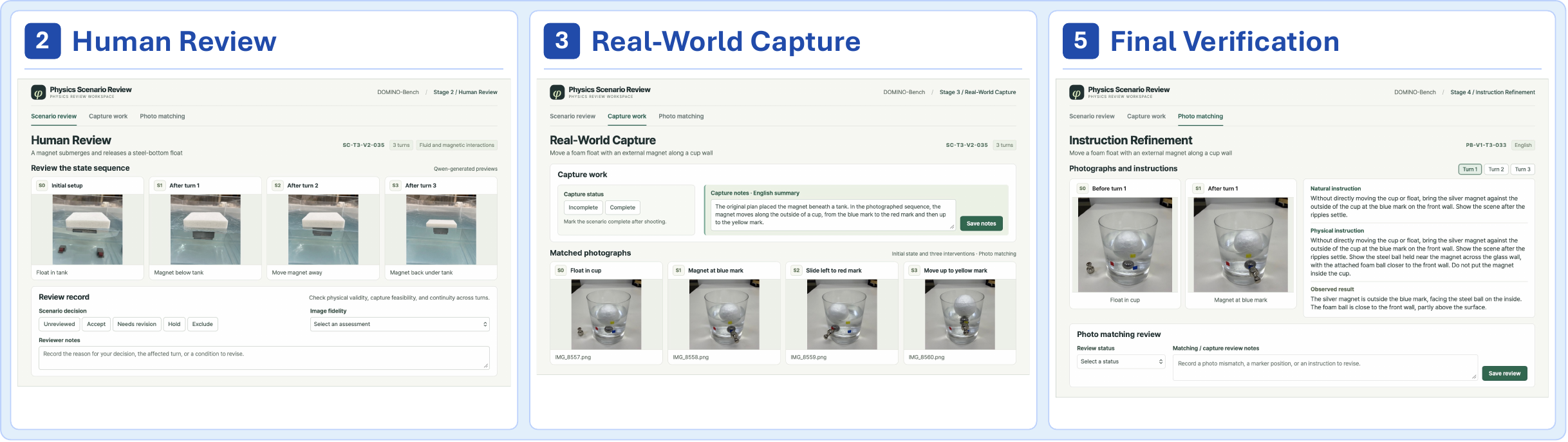}
    \caption{\jh{
        \textbf{Human review interfaces for \benchmarkname{} construction.}
        From left to right, the panels correspond to
        Stages~2, 3, and~5 of the pipeline in
        Fig.~\ref{fig:benchmark_pipeline}.
        \textbf{Left (Stage~2):} Reviewers assess scenario validity,
        capture feasibility, and state continuity, and record
        review decisions and revision notes.
        \textbf{Middle (Stage~3):} Reviewers document changes made
        during real-world capture and align scenario descriptions
        with the captured photographs.
        \textbf{Right (Stage~5):} Reviewers compare the paired natural
        and physical instructions with the corresponding photographs
        to verify image--text alignment and state consistency.
    }}
    \label{fig:human_verify}
\end{figure*}


\section{Related Work}
\label{sec:supp_related_work}

\subsection{Physically Plausible Image Editing}
\label{sec:rw:physically}

Instruction-guided editors such as InstructPix2Pix~\cite{brooks2023instructpix2pix}
learn to map an input image and an editing instruction to an edited image.
Video provides additional supervision for how physical states change
during an edit. PICA-100K~\cite{pu2026picabench} derives editing pairs
from synthetic videos. PhysicEdit~\cite{zhao2026from} uses intermediate
video frames to train transition queries that guide the editor alongside
textual reasoning about the intervention. ChronoEdit~\cite{wu2026chronoedit}
treats the input and edited images as the first and last video frames
and jointly denoises the target frame with temporal reasoning tokens.
PhysEdit~\cite{wan2026physedit} models an edit as a sequence of
intermediate states and uses physics-aware reinforcement learning.
In \modelname{}, each new instruction acts on the scene established
by earlier edits. The agent uses the current scene graph to infer
physical consequences across interacting entities and specify which
attributes to update and which to preserve.

\subsection{Agentic and Graph-Guided Image Editing}
\label{sec:rw:agentic}

Image editing agents differ in what they plan and when they review
an edit. IMAGAgent~\cite{shen2026imagagent} decomposes instructions
into operations, selects editing tools, and uses feedback on edited
images to refine its results. ReasonEdit~\cite{yin2026reasonedit}
interprets instructions before editing and reviews generated images
for unintended changes. RePlan~\cite{qu2026replan} grounds planned
operations in target regions and uses those regions to guide attention
during editing.

Scene graphs make the entities and relations involved in an edit
explicit. SGEdit~\cite{SGEdit} learns tokens for individual objects
and uses an LLM controller to guide an attention-modulated diffusion
editor. VENUS~\cite{VENUS} conditions the target object and its
background separately and uses noise inversion to preserve unedited
regions. In \modelname{}, the Sentinel reviews the target graph and
regional editing conditions before image editing. CRA then guides a
frozen editor with update and preservation clauses. Both types of
clauses can apply to the same region when an edit changes one
attribute of an entity and preserves another.

\subsection{Benchmarks for Physically Plausible Image Editing}
\label{sec:rw:benchmarks}

MagicBrush~\cite{zhang2023magicbrush} provides manually annotated
source images, instructions, and target images for single- and
multi-turn editing. IV-Edit~\cite{qu2026replan} tests fine-grained
grounding and edits that require world knowledge.
PBench-Edit~\cite{wu2026chronoedit} uses image--prompt pairs that
require physical consistency. PICABench~\cite{pu2026picabench}
assesses the physical consequences of an edit through PICAEval,
which uses human annotations to identify relevant regions and asks
a vision-language model questions about those regions.
PhyEditBench~\cite{guo2026phyeditbench} extracts successive states
from real videos and evaluates edits between stages of a physical
process. \benchmarkname{} follows distinct interventions applied
successively to the same scene. Each turn after the first starts
from the editor's preceding output. Photographs taken before and
after each intervention provide references for how a later action
changes earlier physical states and which attributes it preserves.

\section{Benchmark Details}
\label{sec:supp_benchmark_details}

\subsection{Benchmark Construction Pipeline}
\label{sec:supp_benchmark_construction}

\benchmarkname{} evaluates physical plausibility and state
consistency across successive image edits.
\jh{It covers five physical domains: rigid-body mechanics, deformation
and fracture, fluid interactions, optics, and magnetic interactions.
It includes single- and mixed-physics sequences of up to three turns
and five- and ten-turn sequences for long-horizon evaluation.}

\paragraph{\jh{Stage 1: Scenario generation.}}
To turn these evaluation requirements into scenarios for real-world
capture, GPT-6 drafts scenarios in Stage~1 of
Fig.~\ref{fig:benchmark_pipeline}.
The generation follows three design criteria: multi-turn sequences,
mixed-physics scenarios, and cause-and-effect continuity.
See Fig.~\ref{fig:domino_bench_construction_prompts} (Stage~1)
for a condensed version of the scenario-generation prompt.

\paragraph{\jh{Stage 2: Human review.}}
\jh{Human reviewers assess physical validity, capture feasibility,
and state continuity across turns.}
Qwen-Image generates preview images solely to help reviewers inspect
the proposed scenes and changes.
Scenarios requiring revision are updated using review feedback
and reviewed again before capture.
See Fig.~\ref{fig:human_verify} (Stage~2) for the human
review interface.
\jh{Across 14 batches, the review database records 549 distinct
batch--scenario pairs: 166 accepted (30.2\%), 236 excluded (43.0\%),
78 marked for revision (14.2\%), 68 on hold (12.4\%), and 1 unreviewed
(0.2\%). These figures describe recorded review statuses, not first-pass
acceptance or retention into the final benchmark.}

\begin{figure}[!htbp]
    \centering
    \includegraphics[width=0.9\columnwidth]{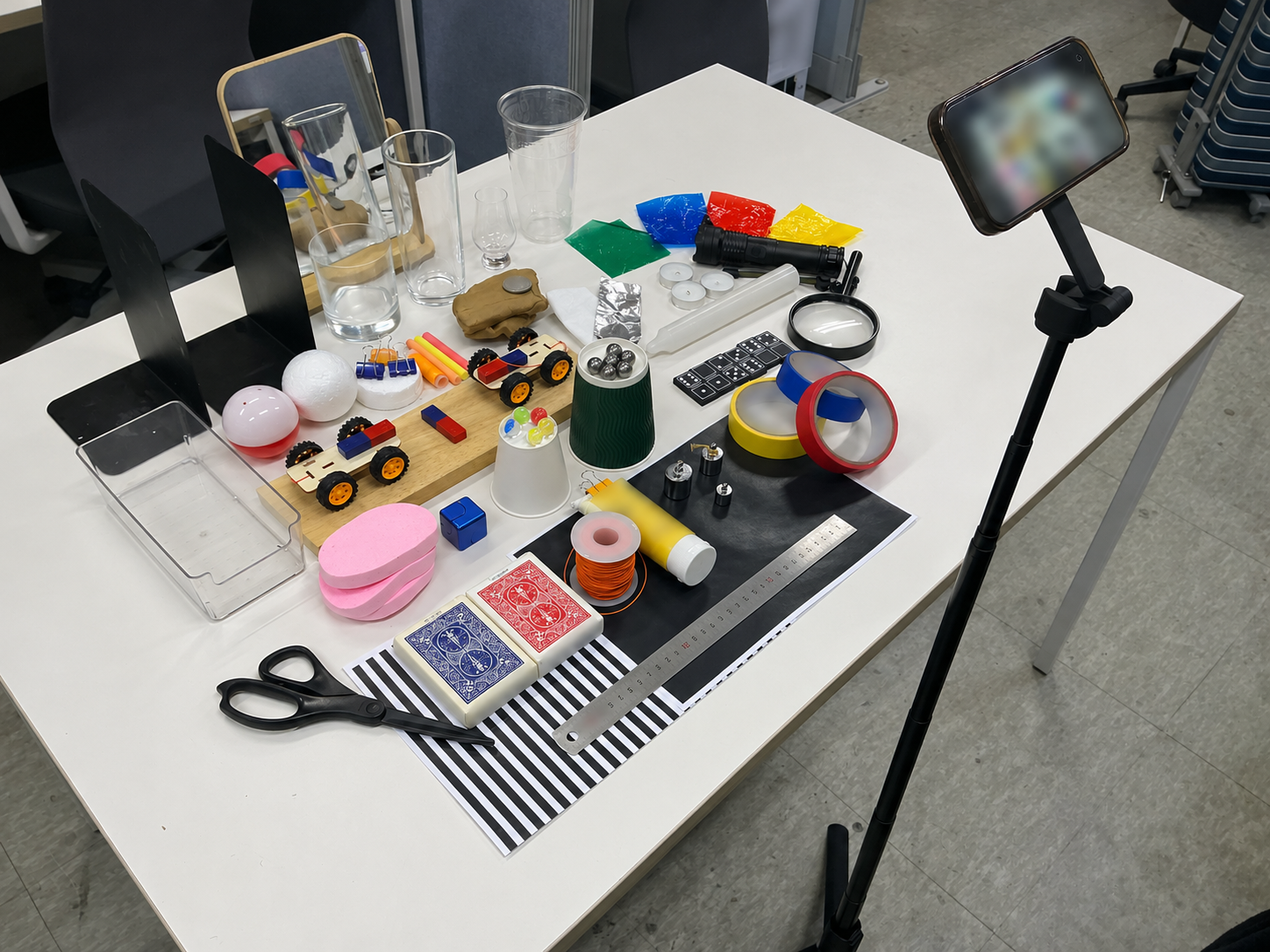}
    \caption{
        Photograph of the capture environment and props
        used to construct \benchmarkname{}.
    }
    \label{fig:capture_env}
\end{figure}

\paragraph{\jh{Stage 3: Real-world capture.}}
In Stage~3, we carry out the reviewed scenarios with real objects
to obtain observed references for their physical consequences.
Figure~\ref{fig:capture_env} shows the environment and props
used to construct \benchmarkname{}.
\jh{We keep lighting consistent and block other illumination only
in scenarios involving light sources.}
An $N$-turn sequence yields $N+1$ photographs of the initial state
and the state after each action.

\jh{Human reviewers revise the scenario descriptions to match the
captured objects, conditions, and observed consequences.}
See Fig.~\ref{fig:human_verify} (Stage~3) for the capture
documentation interface.
Preview images are not used after real-world capture; only the
captured photographs serve as reference images in \benchmarkname{}.

\paragraph{\jh{Stage 4: Instruction refinement.}}
\jh{Using the captured photographs and revised descriptions, GPT-6
refines one natural--physical instruction pair per transition.}
\jh{Natural instructions express everyday editing requests without
specifying their physical consequences and serve as model inputs.
Physical instructions describe observed consequences and states to
preserve and serve as evaluation references.}
Figure~\ref{fig:domino_bench_construction_prompts} (Stage~4)
presents a condensed version of the instruction-refinement prompt
and an example of the paired natural and physical instructions.

\paragraph{\jh{Stage 5: Final verification.}}
Before a sequence is included in the benchmark, human reviewers
compare both instruction types with the corresponding captured
images in Stage~5.
This final verification checks image--text alignment and state
consistency.
See Fig.~\ref{fig:human_verify} (Stage~5) for the final
verification interface.
\jh{The final benchmark contains 151 sequences: 50 one-turn,
45 two-turn, 36 three-turn, 10 five-turn, and 10 ten-turn sequences.}

\section{Additional Examples from DOMINO-Bench}
\label{sec:supp_benchmark_examples}

\begin{figure*}[!tp]
    \centering
    \includegraphics[
        width=\textwidth,
        height=0.79\textheight,
        keepaspectratio
    ]{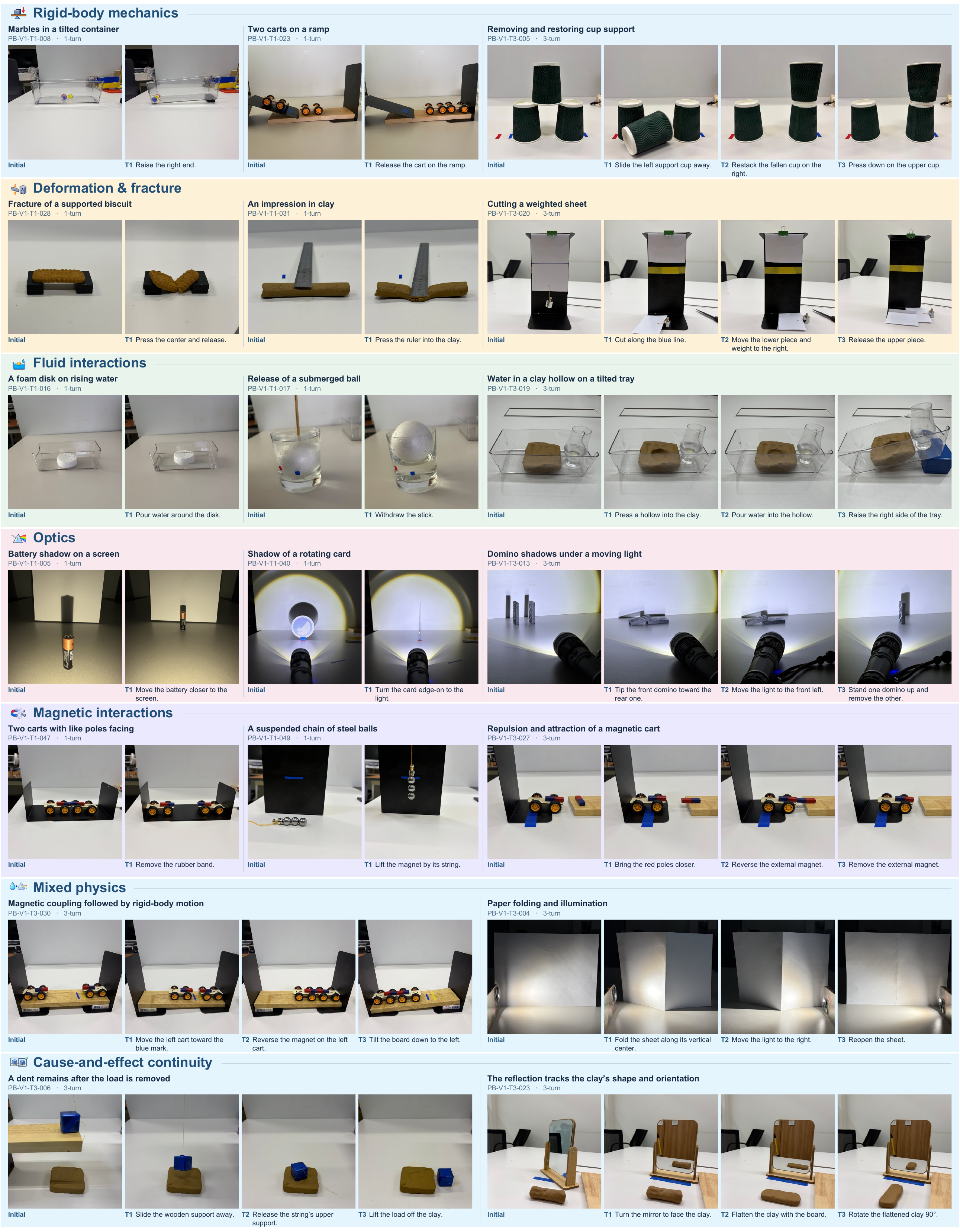}
    \caption{
        \textbf{Additional examples from DOMINO-Bench.}
        Each sequence starts with a photograph of the initial scene,
        followed by the photographed outcome of each edit.
        The instruction appears beneath the corresponding outcome.
        The first five rows each show two one-turn sequences and one
        three-turn sequence for a physical domain: rigid-body mechanics,
        deformation and fracture, fluid interactions, optics, and
        magnetic interactions.
        The sixth row shows how multiple physical domains interact
        within a sequence. The final row illustrates cause-and-effect
        continuity: a dent remains after a load is removed, and a mirror
        reflects the changing shape and orientation of a piece of clay.
    }
    \label{fig:suppl_benchmark_examples}
\end{figure*}

\jh{Fig.~\ref{fig:suppl_benchmark_examples} presents additional examples from DOMINO-Bench covering the five physical domains described in Sec.~\ref{sec:benchmark}: rigid-body mechanics, deformation and fracture, fluid interactions, optics, and magnetic interactions. Beyond individual domains, the figure includes mixed-physics scenarios involving physical effects from multiple domains within a single editing sequence. It also illustrates cause-and-effect continuity, where physical changes introduced in earlier turns establish conditions that influence subsequent outcomes. These examples highlight both the physical consequences of individual interventions and the dependencies that must be maintained across successive edits.}

\begin{figure*}[p]
\begingroup
\definecolor{DominoPromptTeal}{HTML}{32665A}
\definecolor{DominoPromptBlue}{HTML}{405E83}
\newcommand{\DOMINOpromptpanel}[3]{%
  \par\noindent
  \colorbox{#1!14!white}{%
    \begin{minipage}{\dimexpr\linewidth-2\fboxsep\relax}
      \color{#1!75!black}\normalsize\bfseries #2
    \end{minipage}}%
  \par\nointerlineskip\noindent
  \colorbox{#1!4!white}{%
    \begin{minipage}{\dimexpr\linewidth-2\fboxsep\relax}
      \normalsize\setlength{\parskip}{3pt}\setlength{\parindent}{0pt}%
      #3
    \end{minipage}}%
  \par}
\setlength{\fboxsep}{6pt}
\DOMINOpromptpanel{DominoPromptTeal}
  {Stage 1 \enspace\textbar\enspace Scenario Generation}{%
\textbf{Initial scene and interventions.} Design a physically plausible editing scenario that can be photographed with everyday objects. For a sequence of \textit{N} turns, describe the initial arrangement, including relevant contact or support relations, and specify one intervention per turn. For each turn, identify the object to manipulate, the action, when to photograph the result, and the visible change expected after the action. Include indirect effects on other objects or on the scene.

\textbf{Cause-and-effect continuity.} Start each turn from the state left by the preceding turn. Identify which earlier changes persist, which are reversed by the next action, and which affect a later outcome. To test a proposed dependency, ask how the same later action would play out if the earlier change had not occurred. For example, if inserting an object makes water overflow, removing it later need not restore the original water level. Revise sequences that require an unmentioned reset or whose turns could be performed independently with the same results.

\textbf{Physics coverage and candidate selection.} Draw from rigid-body mechanics, deformation and fracture, fluid interactions, optics, and magnetic interactions. In a mixed-physics scenario, connect effects from different domains through a shared object or state. Prefer candidates with distinct causal mechanisms over variants that only change an object's appearance.

\textbf{Capture and verification plan.} For each selected candidate, separate the user's edit request, the photographer's procedure, and the predicted consequences. State what a reviewer could check in the photograph: whether the action was performed, which new physical effect is visible, and which earlier state persists. Specify a viewpoint or observation time if it matters. Flag conditions that need a trial capture and allow for variation in outcomes such as fracture patterns or spill boundaries.
}
\vspace{2.2mm}
\DOMINOpromptpanel{DominoPromptBlue}
  {Stage 4 \enspace\textbar\enspace Instruction Refinement}{%
\textbf{Match actions to photographed states.} Inspect the full sequence of \textit{N}+1 photographs alongside the reviewed scenario and capture notes. Check their order using the objects, their arrangement, and the changes between adjacent states. Use the photographs to establish what is visible and the capture notes to identify actions performed between exposures. If a prop, position, or action changed during the shoot, describe the photographed sequence and record the deviation.

\textbf{Natural instruction.} For each transition, write an everyday editing request that a reader can follow from the input photograph alone. Identify the action, target, direction or destination, and any necessary constraints or observation time. Use visible landmarks when they resolve ambiguity. Leave the resulting physical changes implicit so that the editor must infer them.

\textbf{Physical instruction.} Keep the same action, target, and constraints. Add the visible consequences in the next photograph and any effects from earlier turns that still hold, such as a crease, a wet patch, or a missing support. Use consistent object names throughout. Record the observed result separately. If a cause or motion path cannot be established from the photographs and notes, describe the visible end state and flag the uncertainty for review.

\textbf{Example from the magnet-and-float sequence (Turn 1).}
\par\emph{Natural.} Bring the silver magnet to the blue mark on the outside of the cup without moving the cup or float directly. Show the scene after the ripples settle.
\par\emph{Physical.} Bring the silver magnet to the blue mark on the outside of the cup without moving the cup or float directly. Show the scene after the ripples settle. Show the steel ball inside the cup next to the magnet across the glass, with the attached foam float near the front wall.
}
\endgroup
\caption{\textbf{GPT prompts for DOMINO-Bench construction.}
Condensed English versions of the instructions used to draft scenarios
(Stage~1) and to write paired editing instructions from real photographs
(Stage~4). The example illustrates how both instructions specify the same
intervention while the physical instruction also describes its visible outcome.}
\label{fig:domino_bench_construction_prompts}
\end{figure*}

\clearpage


\section{Method Details}
\label{sec:supp_method_details}

\subsection{Overall Algorithm and Agent Prompts}
\label{sec:supp_algorithm_prompts}

Algorithm~\ref{alg:supp_domino} traces the edit from the current
image $I_t$ to the next image $I_{t+1}$. The Weaver, Forecaster,
and Sentinel share a frozen Qwen3-VL-8B-Instruct and use separate
calls. Once per turn, the Forecaster infers physical consequences
and records them in a causal specification. If the Sentinel identifies a defect, the Forecaster revises
the candidate against that same specification. The plan review
and final compilation are detailed in
Secs.~\ref{sec:supp_plan_review} and~\ref{sec:supp_graph_to_region}.

\begin{algorithm}[H]
\caption{Cause-and-effect-aware multi-turn editing}
\label{alg:supp_domino}
\footnotesize
\textbf{Input} Initial image $I_1$ and natural instructions
$u_1,\ldots,u_T$.\par
\textbf{Output} Edited images $I_2,\ldots,I_{T+1}$ for completed turns.
\par\smallskip
\begingroup
\newcommand{\DStep}[2]{\item\hspace*{#1em}#2}
\begin{list}{\scriptsize\arabic{enumi}}{%
  \usecounter{enumi}\setlength{\leftmargin}{1.75em}%
  \setlength{\labelwidth}{1.25em}\setlength{\labelsep}{.5em}%
  \setlength{\itemsep}{.5pt}\setlength{\parsep}{0pt}%
  \setlength{\topsep}{0pt}\setlength{\partopsep}{0pt}}
\DStep{0}{\textbf{for} $t=1,\ldots,T$ \textbf{do}}
\DStep{1}{$G_t \gets \textsc{Weaver}(I_t,u_t)$ (up to 5 attempts, Fig.~\ref{fig:model}, block~1)}
\DStep{1}{\textbf{if} no valid scene graph \textbf{then stop}}
\DStep{1}{$S \gets \textsc{ForecasterInfer}(I_t,u_t,G_t)$ (Fig.~\ref{fig:model}, block~2)}
\DStep{1}{Initialize revision history $H$ and executable plan $L$ to empty.}
\DStep{1}{\textbf{for} attempt $=1,\ldots,5$ \textbf{do}}
\DStep{2}{$C \gets \textsc{ForecasterPlan}(I_t,u_t,G_t,S,H)$ (Fig.~\ref{fig:model}, block~2)}
\DStep{2}{Compile clauses and regions $R$ from $C$ (Fig.~\ref{fig:model}, block~4).}
\DStep{2}{\textbf{if} invalid \textbf{then} save $C$ and error in $H$ and \textbf{continue}}
\DStep{2}{\textbf{if} graph and controls repeat and $L$ exists \textbf{then break}}
\DStep{2}{$V \gets \textsc{SentinelReview}(I_t,u_t,G_t,S,C,R)$ (Fig.~\ref{fig:model}, block~3)}
\DStep{2}{Set $L\gets(C,R,V)$ and add the candidate and findings to $H$.}
\DStep{2}{\textbf{if} review $V$ and contradiction checks pass \textbf{then break}}
\DStep{1}{\textbf{end for}}
\DStep{1}{\textbf{if} $L$ is empty \textbf{then stop}}
\DStep{1}{Compile the final prompt $P_t$ and regional editing conditions $R_t$ from $L$ (Fig.~\ref{fig:model}, block~4).}
\DStep{1}{\textbf{if} no regional condition has positive weight \textbf{then stop}}
\DStep{1}{$I_{t+1} \gets \textsc{FrozenEditor}(I_t,P_t,\mathrm{CRA}(R_t))$ (Fig.~\ref{fig:model}, block~4, with CRA bias from Eq.~\ref{eq:cra_bias}).}
\DStep{0}{\textbf{end for}}
\end{list}
\endgroup
\end{algorithm}

Fig.~\ref{fig:supp_algorithm_prompts} shows the prompts for scene graph construction, physical consequence inference, editing plan construction and refinement, and plan review.

\begin{figure*}[tp]
\begingroup
\definecolor{DominoPromptTeal}{HTML}{32665A}
\definecolor{DominoPromptBlue}{HTML}{405E83}
\newcommand{\DOMINOpromptpanel}[3]{%
  \par\noindent
  \colorbox{#1!14!white}{%
    \begin{minipage}{\dimexpr\linewidth-2\fboxsep\relax}
      \color{#1!75!black}\normalsize\bfseries #2
    \end{minipage}}%
  \par\nointerlineskip\noindent
  \colorbox{#1!4!white}{%
    \begin{minipage}{\dimexpr\linewidth-2\fboxsep\relax}
      \normalsize\setlength{\parskip}{3pt}\setlength{\parindent}{0pt}%
      #3
    \end{minipage}}%
  \par}
\setlength{\fboxsep}{5pt}
\DOMINOpromptpanel{DominoPromptTeal}
{Weaver \enspace\textbar\enspace Construct the current scene graph}{%
\textbf{Observe the source.} Inspect the current image. Use the instruction
to identify the entities, supporting surfaces, and visible interaction
effects relevant to the requested edit. Record the state before the
intervention. Distinguish a visible contact from evidence of support.

\textbf{Return the graph.} Give each entity an ID, kind, label, visibility,
and image box. Record observed appearance, material, and physical state
where relevant. For each interaction, give the source ID, relation, and
target ID. Localize a visible effect with its own region. Use a null
region when its extent cannot be established. Return the
\texttt{nodes} and \texttt{edges} as one JSON graph.}
\vspace{2.2mm}
\DOMINOpromptpanel{DominoPromptBlue}
{Forecaster \enspace\textbar\enspace Infer physical consequences}{%
\textbf{Trace the intervention.} Use the image, instruction, and current
graph. Describe the initial conditions that matter, the action, and
the immediate visible outcome, including changes to indirectly affected
entities. Use the requested observation time when given.

\textbf{Return the causal specification.} For each dependency, give an
item ID, the affected entities and property, and the \texttt{before},
\texttt{cause}, and \texttt{after} descriptions. Identify any specific
property that must be preserved. Summarize the resulting scene in
\texttt{terminal\_scene} and record conflicts or uncertainties.}
\vspace{2.2mm}
\DOMINOpromptpanel{DominoPromptBlue}
{Forecaster \enspace\textbar\enspace Construct or refine the editing plan}{%
\textbf{Implement the specification.} Construct the complete target graph.
Retain IDs for surviving entities. Encode each justified consequence
in the affected attributes, boxes, or interactions. For example, a
change of support can require a new object position and contact relation
even when identity and material are preserved.

\textbf{Return the candidate.} Provide \texttt{target\_graph},
\texttt{requirements}, and \texttt{preserve\_constraints}. In
\texttt{causal\_implementation}, map each specification item to the node
or edge IDs that realize it. Record any unresolved conflict. On a
revision, use the previous candidate and review feedback to correct
the affected entries while preserving earlier correct changes.}
\vspace{2.2mm}
\DOMINOpromptpanel{DominoPromptTeal}
{Sentinel \enspace\textbar\enspace Review the candidate before editing}{%
\textbf{Check the evidence and the plan.} Compare the causal specification
with the source image and instruction. Then inspect the actual target
graph, graph differences, editing clauses, and assigned regions. Check
whether they implement the required physical state updates and preserve
the specified attributes.

\textbf{Return actionable feedback.} For each demonstrated defect,
name the affected graph IDs, describe the evidence, and specify which
outcome and fields must change. Record unresolved evidence under
\texttt{uncertainties}. Return the verdict, issues, defects,
uncertainties, and a short summary. Set \texttt{passed} to true only
when all three lists are empty.}
\endgroup
\caption{\textbf{Prompts for cause-and-effect-aware planning.}
The panels condense the task instructions and output fields for the
four call types. The two Forecaster calls separate inferring the
physical consequences from implementing them in the target graph.
A revision uses the same causal specification and the preceding review.}
\label{fig:supp_algorithm_prompts}
\end{figure*}

\subsection{Scene Graph Representation}
\label{sec:supp_scene_graph}

At each turn, the Weaver extracts $G_t=(\mathcal{V}_t,\mathcal{E}_t)$
from the current image $I_t$ and instruction $u_t$.
The instruction identifies what to inspect. The graph records
the scene before the requested edit. From the second turn onward,
the Weaver observes the preceding edited image and constructs a fresh
graph. Within a turn, entities shared by $G_t$ and its target graph
$G_t^*$ retain their IDs.

\paragraph{Entity nodes.}
Each node records an ID, a descriptive label, a kind
(object, part, substance, surface, or light), visibility
(full, partial, hidden, or unknown), and a \texttt{bbox} field.
The optional \texttt{appearance} and \texttt{state} fields accept
named attributes, such as color and wetness. Material and physical
tags can supply further properties, such as whether an entity
is movable or opaque. Missing attribute dictionaries, material
strings, and tag lists default to empty values.

\paragraph{Interaction edges and image regions.}
A directed edge records its own ID, the source and target node IDs,
and a relation type, such as \texttt{CONTACTS}, \texttt{SUPPORTS},
or \texttt{CASTS\_SHADOW\_ON}. Optional fields describe how the
entities interact (\texttt{mode}), the appearance of the effect,
and its image region. A node's \texttt{bbox} localizes the entity.
An edge's \texttt{region} localizes the effect on the receiving
entity or at the interaction site. Both use
$[x_{\min},y_{\min},x_{\max},y_{\max}]$ with the origin at the upper
left and each axis scaled to $[0,1000]$ over the full image.
A non-null box must have positive area. Unlocalized geometry is
recorded as null. For example, a support edge can connect two
entities without a separate region.

\begin{figure*}[t]
    \centering
    \includegraphics[width=\textwidth]{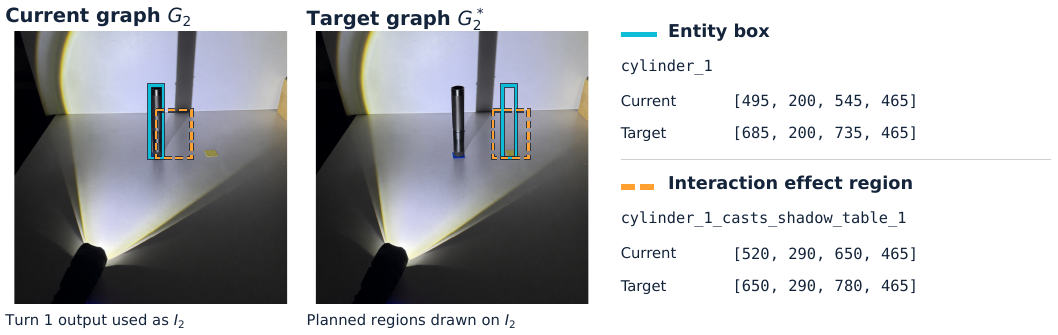}
    \caption{\textbf{Entity and interaction regions for a cylinder relocation.}
    Both overlays use the actual Turn~2 input. The current graph
    locates the cylinder and its cast shadow, and the reviewed target
    graph specifies where each should move. The coordinates are the
    Weaver's and Forecaster's graph estimates in the full-image frame.}
    \label{fig:supp_scene_graph}
\end{figure*}

\paragraph{Moving a cylinder under fixed illumination.}
In Fig.~\ref{fig:supp_scene_graph}, the edited image from
Turn~1 serves as the input $I_2$ for Turn~2.
The instruction asks the editor to move the upright cylinder's
base from the blue mark to the yellow tape while keeping the
flashlight fixed.
The cylinder node's box locates the cylinder, while the shadow
edge's region locates its cast shadow on the table.
The target graph shifts both regions and preserves the
flashlight's box.

\subsection{Plan Review and Refinement}
\label{sec:supp_plan_review}

\paragraph{What the Sentinel reviews.}
The Sentinel uses a separate call to the same frozen VLM. It first
checks whether the instruction and current image support the Forecaster's
causal specification. It then compares the current and target graphs,
their differences, and the proposed editing clauses and regions.
For instruction following, it checks whether the requested intervention
acts on the intended entity. For physical plausibility, it checks
whether the target graph implements the inferred physical consequences
and whether the clauses agree with the corresponding graph fields.
For preservation, it checks whether an attribute marked for preservation
conflicts with a required update. The review takes place before rendering
and assesses the proposed scene.

\paragraph{Feedback and stopping.}
The Sentinel returns a verdict, issues, defects, uncertainties, and
a short summary. Each defect identifies the affected graph IDs when
available, describes the discrepancy, and specifies a required change.
Uncertain source evidence is recorded separately. A candidate passes
when the reviewer approves it, the three lists are empty, and the
target graph passes the implementation's contradiction checks.
Otherwise the Forecaster receives the previous candidate and feedback
to revise the graph and editing conditions against the same causal
specification. Planning stops at the first pass, when an executable
candidate repeats the same target graph and controls, or after five
candidate attempts. In the latter two cases, the planner can select
the latest executable graph with its previous review status and
unresolved findings. If no graph is executable, or its final
compilation yields no positive-weight attention target, the edit fails.

\paragraph{What changed after review.}
In Turn~2 of the cylinder example (Fig.~\ref{fig:supp_scene_graph}),
the first candidate described the cylinder on the yellow tape but
kept its original box, its contact with the blue mark, and the original
shadow region. The Sentinel returned two defects. One called for
moving the cylinder and replacing its contact edge. The other called
for shifting the shadow on the table. The Forecaster's second candidate
moved the cylinder box and shadow region to the locations shown in
the figure, replaced the blue-mark contact with a yellow-tape contact,
and preserved the flashlight box. The Sentinel accepted this candidate.
A separate edge still described the blue mark as supporting the cylinder,
a residual inconsistency the review missed.

\subsection{Graph-to-Region Compilation and CRA}
\label{sec:supp_graph_to_region}

\begin{figure*}[t]
    \centering
    \includegraphics[width=\textwidth]{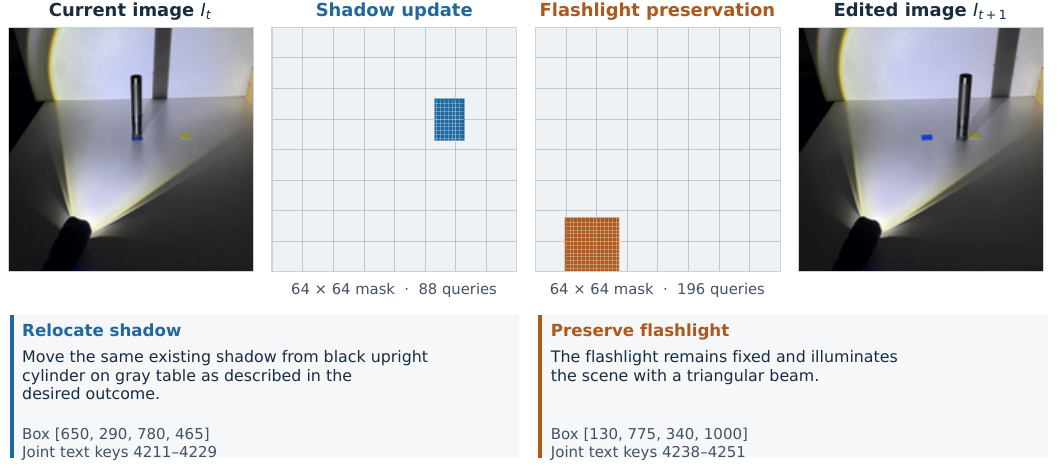}
    \caption{\textbf{From reviewed clauses to spatial guidance in turn~2.}
    The shadow relocation and flashlight preservation clauses select
    different text keys and target-image queries. The masks are
    reconstructed from the executed regions and show which queries
    receive the inside bias. Input and output are from the same turn.}
    \label{fig:supp_cra_visualization}
\end{figure*}

\paragraph{From graph differences to regions.}
At turn $t$, Graph-to-Region Compilation compares the current and
target graphs, $\Delta G_t=G_t\ominus G_t^*$, using the shared IDs.
For an entity, it reads the bounding box from its node. For a
localized interaction effect, such as a cast shadow, it reads the
effect region from the edge. The operation determines which graph
provides the region.
\begin{itemize}
    \item \textbf{Add, modify, or relocate.} An addition uses the
    new entity or effect region in $G_t^*$. A modification uses the
    target region of the affected entity or effect. A relocation
    binds the clause to its destination in $G_t^*$. The global
    instruction still describes the move from the original location.
    \item \textbf{Remove or restore.} A removal uses the footprint
    of the entity or effect in $G_t$. A restoration clause uses the
    source footprint to specify where the previously occupied
    region should be cleared.
    \item \textbf{Preserve.} A grounded preservation clause uses the
    target region of its entity or effect and names the attribute to
    preserve. A clause with several identified entities can cover the
    union of their boxes. An update and a preservation clause can
    therefore apply to the same region while addressing different
    attributes.
\end{itemize}
A clause receives regional guidance when its graph entries and
regions can be resolved. Ambiguous preservation statements remain
global. The Sentinel reviews a preview of the candidate's execution
clauses and regions. After selection, the compiler produces the
final prompt $P_t$ and records each routed clause's character span,
region, and confidence weight. $P_t$ contains a framing instruction,
$u_t$, the immediate visible outcome from the causal specification,
the edit and preservation clauses, and a closing instruction to
preserve unrelated content while allowing affected entities and
effects to respond.

\paragraph{From prompt spans to attention.}
For clause $r_i$, we align its character span in $P_t$ to token
offsets in the actual editor input. We map the matched tokens to
their joint text-key positions $\mathcal{T}_i$ after the editor
expands image-token positions. On the target-image query grid,
$m_i(q)=1$ when a cell center lies inside any box for $r_i$.
If the entire region misses all cell centers, cells with positive
area overlap supply the mask. The compiler leaves the original boxes
unchanged. CRA sums the clause contributions to $\boldsymbol B$ as
in Eq.~\ref{eq:cra_bias}. For one attention head it applies
\begin{equation}
\begin{aligned}
\widetilde{\boldsymbol O}
  &= \boldsymbol O_{\mathrm{native}}
     + (\boldsymbol A_{B}-\boldsymbol A_0)\boldsymbol V,\\
\boldsymbol A_{B}
  &= \operatorname{softmax}_{k}(\boldsymbol L+\boldsymbol B),\qquad
\boldsymbol A_0 = \operatorname{softmax}_{k}(\boldsymbol L).
\end{aligned}
\label{eq:supp_cra_output}
\end{equation}
Both softmax operations use the original padding mask and span all
text, reference-image, and target-image keys. Only target-image
queries receive the correction. The native outputs for text and
reference-image queries are retained.

\paragraph{Worked example and settings.}
In turn~2 of Fig.~\ref{fig:supp_cra_visualization}, the instruction
requests moving the cylinder from the blue mark onto the yellow tape while
keeping the flashlight fixed. One of the five routed clauses moves
the cylinder's cast shadow. Its span $[573,681)$ in $P_t$ maps to
joint text keys 4211--4229 (inclusive, zero-based). Its target edge
region $[650,290,780,465]$, expressed as $[x_1,y_1,x_2,y_2]$ on a
0--1000 scale, selects 88 of the $64\times64$ query cells.
A separate clause preserves the flashlight and its beam. Its box
selects 196 cells. These masks specify where the editor receives
each condition. In the rendered image, the cylinder moves toward
the tape but its base falls short of the planned destination.

The frozen Qwen-Image-2.1 editor uses 40 denoising steps with
guidance scale~1 and key/value caching on \benchmarkname{}.
The input aspect ratio is preserved up to rounding each dimension
to a multiple of 32 at approximately $1024^2$ pixels. This example
renders at $1024\times1024$. CRA runs in all 32 transformer blocks during
the first 16 steps (indices 0--15), using $\alpha=1$ and
$\beta_{\mathrm{in}}=\beta_{\mathrm{out}}=2$. Both example clauses
have $c_i=1$, yielding a logit bias of $+2$ inside and $-2$ outside
their respective regions for the corresponding text keys.

\subsection{Interactive Control}
\label{sec:supp_interactive_control}

A user can inspect the editing plan, select an entity in the scene
graph, and adjust the image region associated with its text clause.
For a clause $r_i$, moving or resizing $R_i$ changes the target-image
queries selected by $m_i(q)$ in Eq.~\ref{eq:cra_bias}. CRA then directs
attention to the same clause at the revised locations. The water
example in Fig.~\ref{fig:interactive_multiturn} shows how a user
selects the water and raises the upper boundary of its box.

Fig.~\ref{fig:supp_interactive_control} shows three recorded settings
for each of two entities. For water, the left, right, and lower
boundaries of the box stay fixed as the upper boundary rises. The
visible water surface rises accordingly, while the glass stays on the
table. For the cube, the box moves right and toward the front edge of
the top card. The cube occupies a different position in each image,
with the card structure still supporting it. These recorded examples
show how the selected regions correspond to visible changes in water
level and cube position.

\noindent\begin{minipage}{\linewidth}
    \centering
    \includegraphics[width=.96\linewidth]{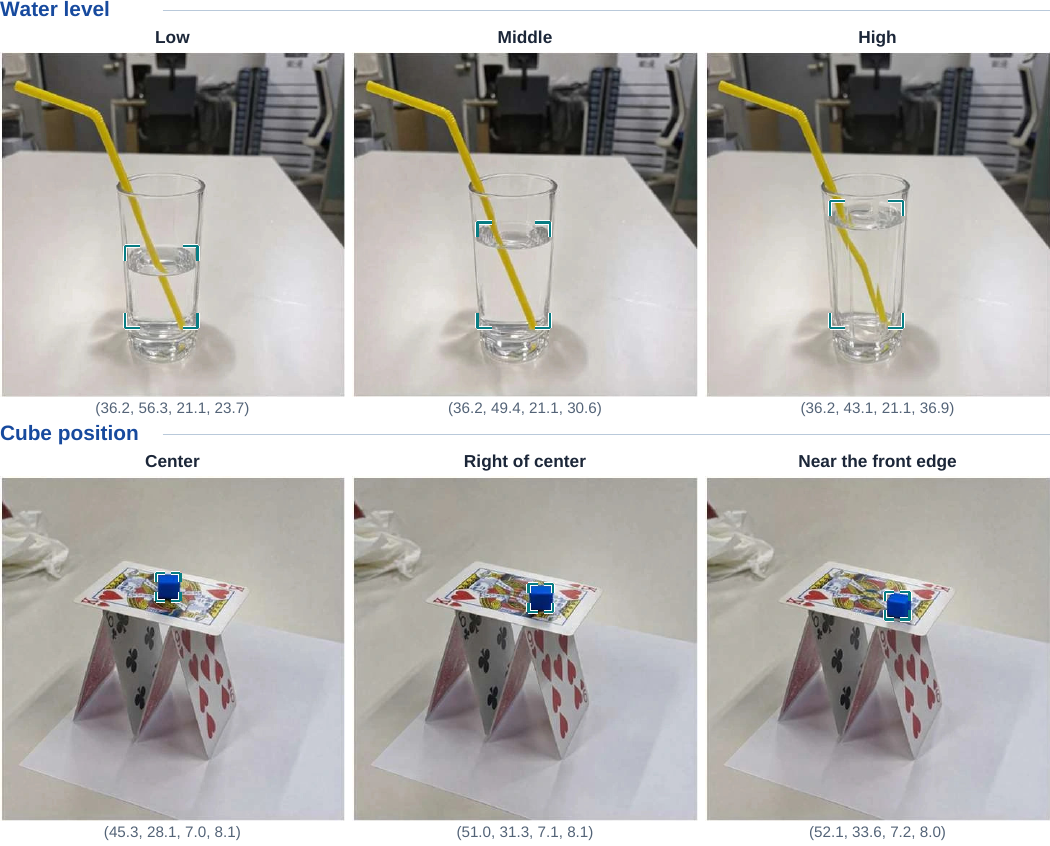}
    \captionof{figure}{\textbf{Interactive control of water level and cube position.}
    Each image shows a recorded setting. Corner marks delineate the
    selected region. The tuple below each image gives its left and top
    coordinates, width, and height as percentages of the displayed
    image dimensions, measured from the upper-left corner.
    Raising the upper water boundary changes the fill level.
    Moving the cube region changes where the cube rests on the card.}
    \label{fig:supp_interactive_control}
\end{minipage}\par

\noindent\textbf{Paired user controls.} 
Fig.~\ref{fig:supp_paired_user_controls} compares two user-selected
settings for each of five additional scenes. Both variants start from
the same image and use the same editor seed and inference settings.
In the mirror scene, the prompt requests the removal of one domino and
its reflection. Moving the two removal regions selects which domino
and reflection disappear. In the color-filter scene, the two regions
continue to cover the gel and the light it casts. Changing the color
in the prompt guides the editor to render red or blue illumination.
The candle, card, and waterline examples change both the target wording
and its associated regions.

\noindent\begin{minipage}{\linewidth}
    \centering
    \includegraphics[width=.96\textwidth]{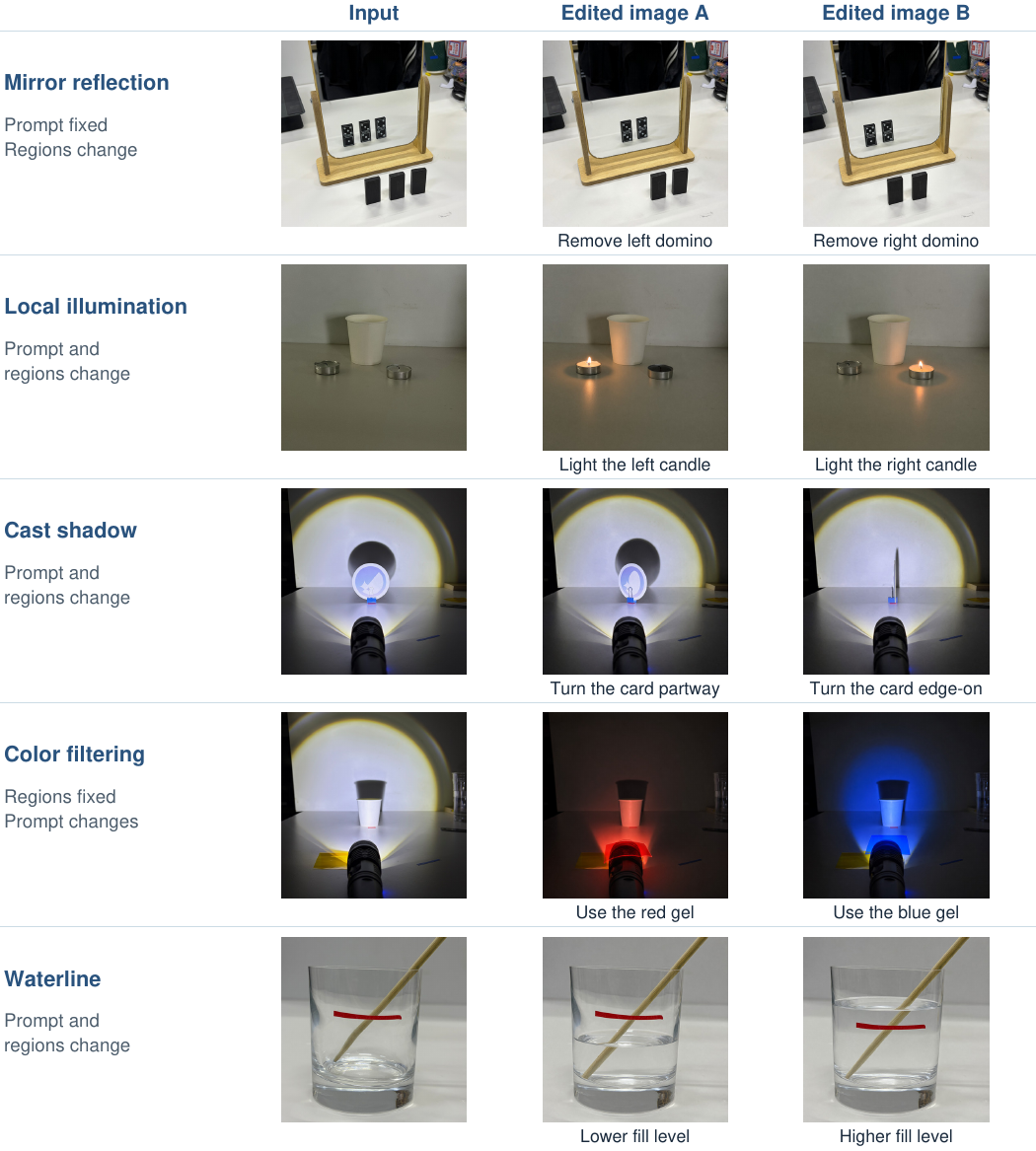}
    \captionof{figure}{\textbf{Paired interactive edits with CRA.}
    Each row pairs one input with two precomputed outputs from the
    frozen editor. The first column identifies which controls differ
    between the two outputs. In the mirror example, the two selected
    regions cover the physical domino and its reflection.}
    \label{fig:supp_paired_user_controls}
\end{minipage}\par

\section{Experimental Settings and Evaluation Protocols}
\label{sec:supp_experimental_details}

\begin{figure*}[t]
    \centering
    \includegraphics[width=.96\textwidth]{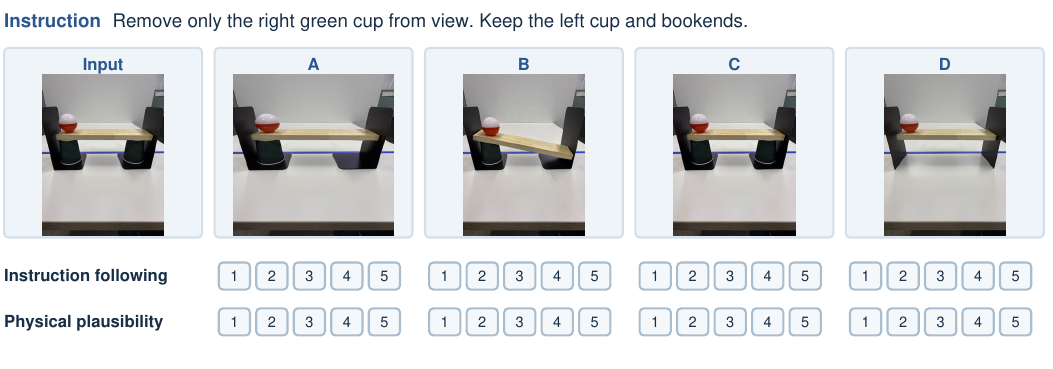}
    \caption{\textbf{Human evaluation protocol.}
    This reconstruction shows an actual study case with its
    input, instruction, and four anonymized outputs in an example
    order. Each output receives a separate five-point rating
    for each criterion.}
    \label{fig:supp_human_evaluation}
\end{figure*}

\subsection{Implementation and Baseline Settings}
\label{sec:supp_implementation}

\paragraph{Models and inference.}
The Weaver, Forecaster, and Sentinel share frozen
Qwen3-VL-8B-Instruct~\cite{yang2025qwen3}.
The Consequence-Guided Editor uses frozen
Qwen-Image-2.1~\cite{teamqwen} in bfloat16.
We preserve the input aspect ratio when setting the output
resolution, scaling to approximately $1024^2$ pixels and
rounding each dimension to a multiple of 32.
The editor takes 40 inference steps with a guidance scale of
1 and uses a key--value cache. CRA applies to the first
16 steps in transformer blocks 0 through 31, with
$\alpha=1$ and $\beta_{\mathrm{in}}=\beta_{\mathrm{out}}=2$
(Sec.~\ref{sec:supp_graph_to_region}).
We set the planner seed to 176358989.
The external-benchmark runs use editor seed 42.

\paragraph{Benchmark inputs.}
On \benchmarkname{}, the first turn begins with the
photographed input and the natural instruction. Each later
turn uses the method's own preceding output. \modelname{}
extracts a fresh scene graph from this image.
The photographed outcomes and physical instructions serve
as evaluation references.
The MagicBrush development split contains 266 editing
sequences with 528 turns. The first turn uses the original
image, and later turns use the previous edited image.
Our MagicBrush runs use the editing instructions without
ground-truth targets or masks.
On PICABench, we use each of the 900 single-turn inputs
with its \emph{superficial prompt}. The corresponding
target annotations are reserved for evaluation.

\paragraph{Comparison and runtime.}
Tab.~\ref{tab:main_comparison} compares image editors
(DreamOmni2 and CoCoEdit), multi-turn editors (FreqEdit and
MTC), physics-aware editors (ChronoEdit and PhysicEdit),
graph-based editors (SGEdit and VENUS), and agent-based
editors (ReasonEdit-S and RePlan). For the Sentinel ablation,
the Weaver, Forecaster, and CRA are fixed. The full method
reviews the plan before rendering, and the image-review
variant reviews the edited image. The percentage reported
in Tab.~\ref{tab:ab_sentinel} divides each variant's mean
runtime by the mean runtime of the full method.

\subsection{Automatic Evaluation}
\label{sec:supp_automatic_evaluation}

\paragraph{Editing and physical evaluation.}
EditReward~\cite{wu2026editreward} (ER) gives a learned
preference score for instruction following and visual
quality. MB Physical uses MultiBanana's Physical Realism
rubric~\cite{oshima2026multibanana}.
KRIS Knowledge uses the Knowledge Plausibility
rubric~\cite{wu2026kris}.
For \benchmarkname{}, a local Qwen3-VL-8B evaluator applies
these two rubrics. The KRIS evaluation uses benchmark
explanations on all three benchmarks.
On \benchmarkname{}, CLIP-I compares the edited image with
the current input, and CLIP-T compares the edited image
with the current instruction~\cite{pmlr-v139-radford21a}.
For a later turn, the current input includes edits made by
the same method in earlier turns.

\subsection{Human Evaluation Protocol and Interface}
\label{sec:supp_human_evaluation}

The study covers 30 cases, with ten each from \benchmarkname{},
MagicBrush, and PICABench. Each case presents the input image,
the editing instruction, and four edited images from
\modelname{}, ReasonEdit-S, FreqEdit, and RePlan.
The interface identifies the outputs by letters A through D.
Participants assign two independent integer scores from
1 to 5 to each output. For \emph{instruction following},
they assess whether the output carries out the requested
edit. For \emph{physical plausibility}, they assess whether
the resulting scene reflects the physical consequences of
the edit. In the case shown in Fig.~\ref{fig:supp_human_evaluation},
the first judgment asks whether the specified cup disappears.
The second asks how the board responds when it loses that support.

The study records each score with its case, session, criterion,
and display letter. The letter-to-model mapping for each case
and session recovers the method for analysis.
Tab.~\ref{tab:user_study} reports the mean ratings from
41 evaluation sessions, separately for each criterion and
benchmark.

\subsection{Comparison of Automatic and Human Evaluation}
\label{sec:supp_auto_human_comparison}

We quantify agreement between automatic scores and human
ratings for the four methods common to
Tabs.~\ref{tab:main_comparison} and~\ref{tab:user_study}.
Within each benchmark, we use Kendall's $\tau_b$~\cite{kendall1945ties} to measure
how closely the automatic scores and human ratings rank
the four methods.
It compares the order of each of the six method pairs
using the reported mean scores and accounts for ties.
We compare EditReward (ER)~\cite{wu2026editreward} with human
ratings of instruction following (IF). We compare MB
Physical~\cite{oshima2026multibanana} and KRIS
Knowledge~\cite{wu2026kris} with human ratings of physical
plausibility (PP).

\begin{table}[t]
    \centering
    \caption{
        \textbf{Rank agreement between automatic scores and
        human ratings.}
        Each entry gives Kendall's $\tau_b$ across four methods
        on one benchmark.
    }
    \label{tab:supp_rank_agreement}
    \small
    \setlength{\tabcolsep}{4pt}
    \begin{tabular}{@{}lccc@{}}
        \toprule
        \textbf{Benchmark}
        & \textbf{ER--IF}
        & \textbf{MB--PP}
        & \textbf{KRIS--PP} \\
        \midrule
        \benchmarkname{} & 0.00 & 0.55 & 0.18 \\
        MagicBrush       & 0.91 & 0.18 & 0.67 \\
        PICABench        & 0.67 & 0.67 & 0.67 \\
        \bottomrule
    \end{tabular}
\end{table}

On \benchmarkname{}, ER and human IF agree on all three method
pairs involving \modelname{} and reverse the order of all three
baseline pairs, yielding $\tau_b=0.00$
(Tab.~\ref{tab:supp_rank_agreement}). On MagicBrush, ER and IF
give similar rankings ($\tau_b=0.91$), while MB Physical and PP
show weaker agreement ($\tau_b=0.18$). On PICABench, each metric
agrees with its corresponding human criterion on five of six
pairs ($\tau_b=0.67$). All three metrics favor RePlan over
ReasonEdit-S, while both human criteria favor ReasonEdit-S.

Each coefficient compares benchmark-level rankings. Human ratings
cover ten selected cases per benchmark, while automatic scores
are averaged over valid outputs for each method and metric.
Across all three benchmarks, human ratings place \modelname{} first
on both criteria, and the automatic metrics place it first or joint
first among the same four methods. The two evaluations thus agree
on \modelname{}'s leading position. Which baseline ranks next
depends on both the benchmark and the evaluation criterion.

\section{Additional Qualitative Results}
\label{sec:supp_qualitative_results}

We compare five cases from each of MagicBrush, PICABench, and
\benchmarkname{}. As in Fig.~\ref{fig:qualitative_comparison}, the
columns show RePlan, ReasonEdit-S, PhysicEdit, FreqEdit, and
\modelname{} in that order. For each multi-turn case, the first row
gives the common input. Reading down a model's column follows its
own edited images through every turn.

\subsection{Physical Consequences across Interacting Entities}
\label{sec:supp_entity_consequences}

\begin{figure*}
    \centering
    \includegraphics[width=1\linewidth]{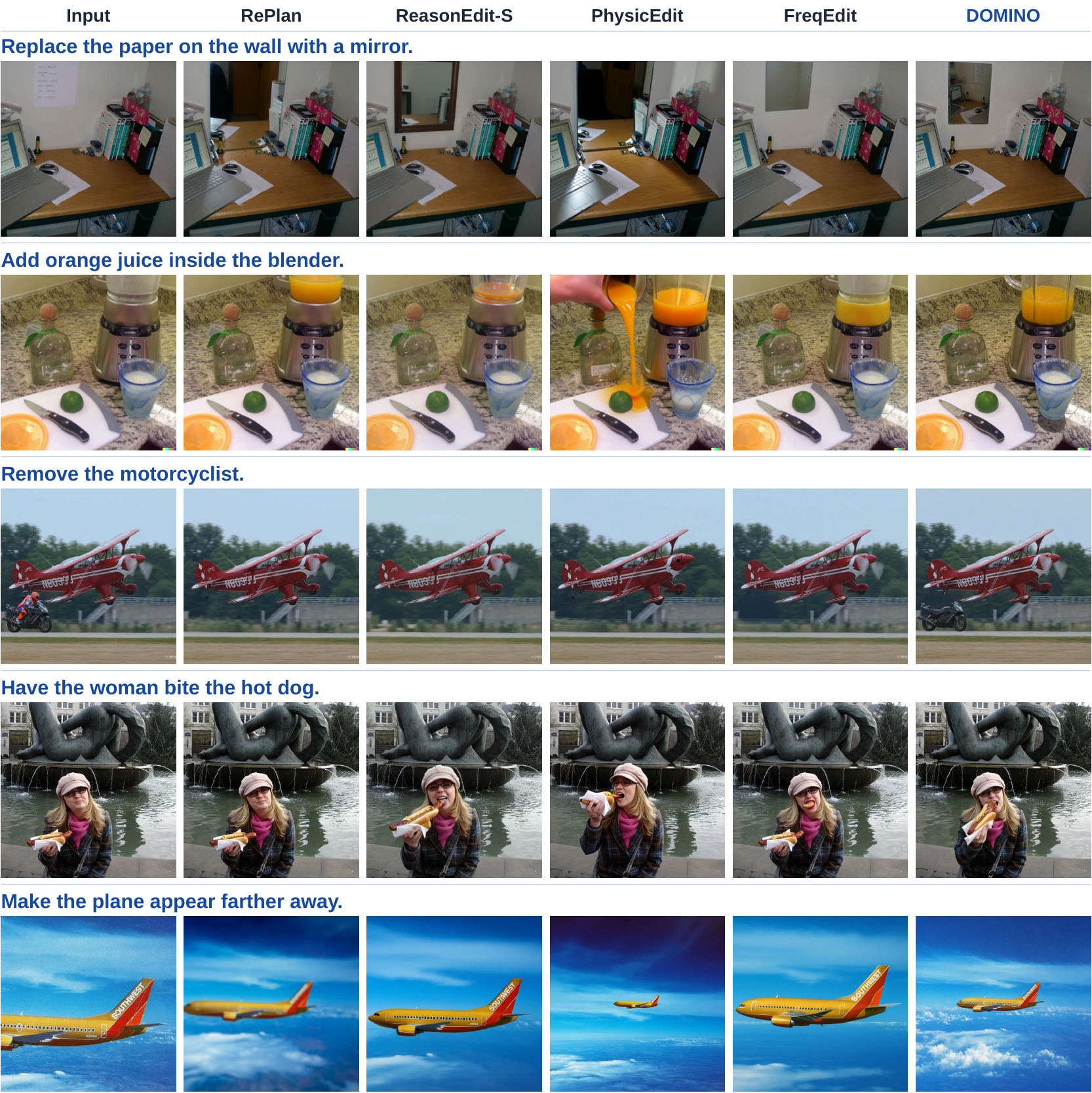}
    \caption{\textbf{Selective updates on MagicBrush.}
    The five instructions change a reflection, add liquid to a container,
    remove a rider, bring food into contact with a person's mouth, and
    change the apparent distance of an aircraft.}
    \label{fig:supp_qualitative_results_1}
\end{figure*}

\begin{figure*}
    \centering
    \includegraphics[width=1\linewidth]{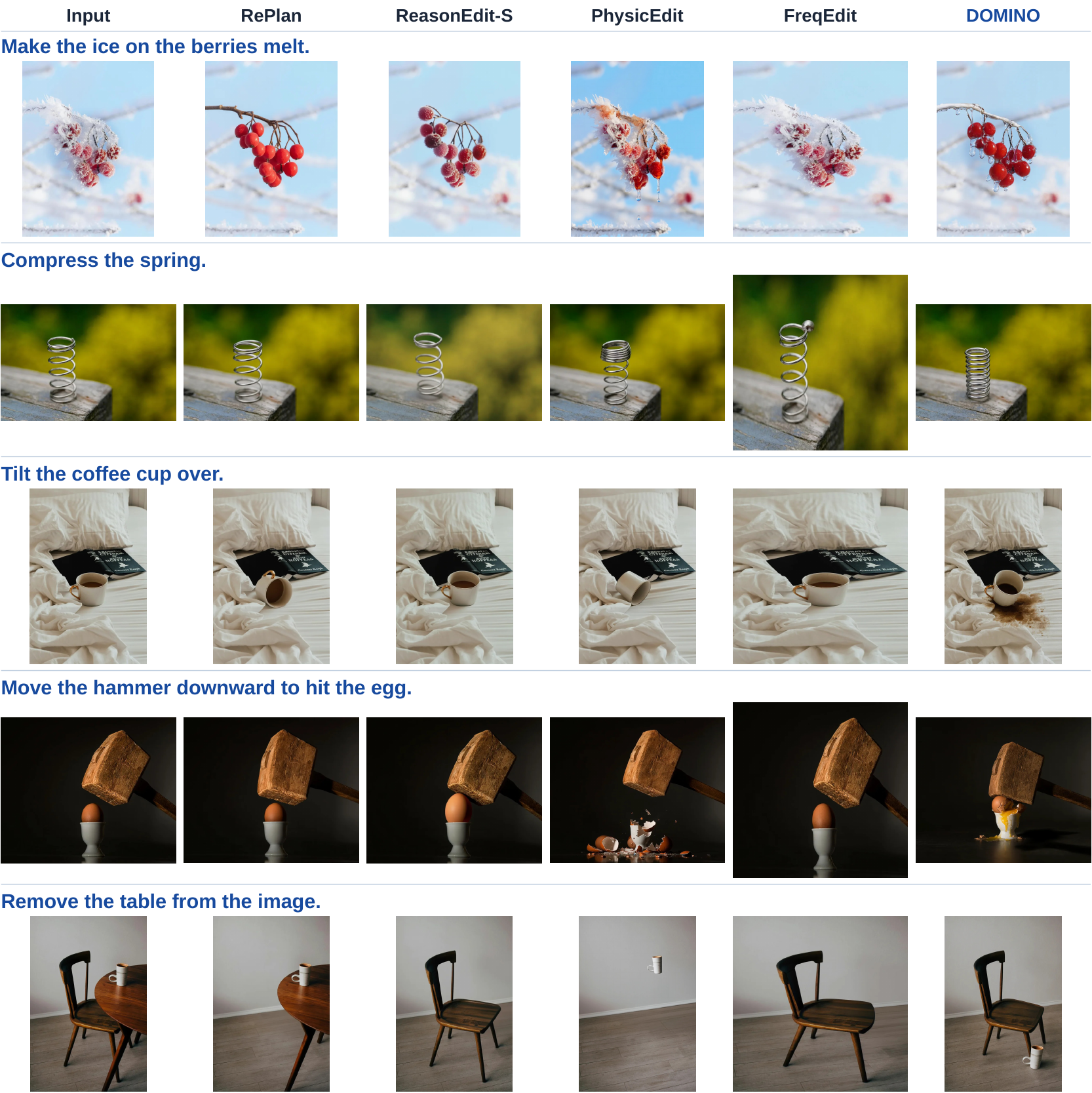}
    \caption{\textbf{Visible physical consequences on PICABench.}
    Melting exposes the berries, compression narrows the gaps between
    coils, and spilled coffee marks the bedding. The last two rows show
    an impact on an egg and the removal of the table supporting a mug.}
    \label{fig:supp_entity_consequences}
\end{figure*}

On MagicBrush (Fig.~\ref{fig:supp_qualitative_results_1}), the
motorcyclist example separates the requested removal from what
should be preserved. All five methods remove the rider, but only
\modelname{} leaves the motorcycle visible in front of the aircraft.
In the desk scene, the mirror in the \modelname{} result reflects
part of the desk, while FreqEdit produces a nearly uniform panel.
In the blender scene, both RePlan and \modelname{} place juice
inside the jar. PhysicEdit also adds a hand pouring juice outside it.
On PICABench (Fig.~\ref{fig:supp_entity_consequences}), tipping the
coffee cup changes both the cup's orientation and the surface beneath it.
RePlan tips the cup. \modelname{} also produces a brown spill on the
bedding. In the support-removal case, the instruction names only the
table. \modelname{} removes it and places the mug on the floor beside
the chair. The mug floats in the PhysicEdit result and disappears in
the ReasonEdit-S and FreqEdit results. The spring tests how far an
update extends within one entity. PhysicEdit compresses the upper
coils, while \modelname{} reduces the gaps along the spring.

\clearpage
\raggedbottom

\twocolumn[{%
    \noindent
    \begin{minipage}{\textwidth}
        \centering
        \includegraphics[width=\linewidth]{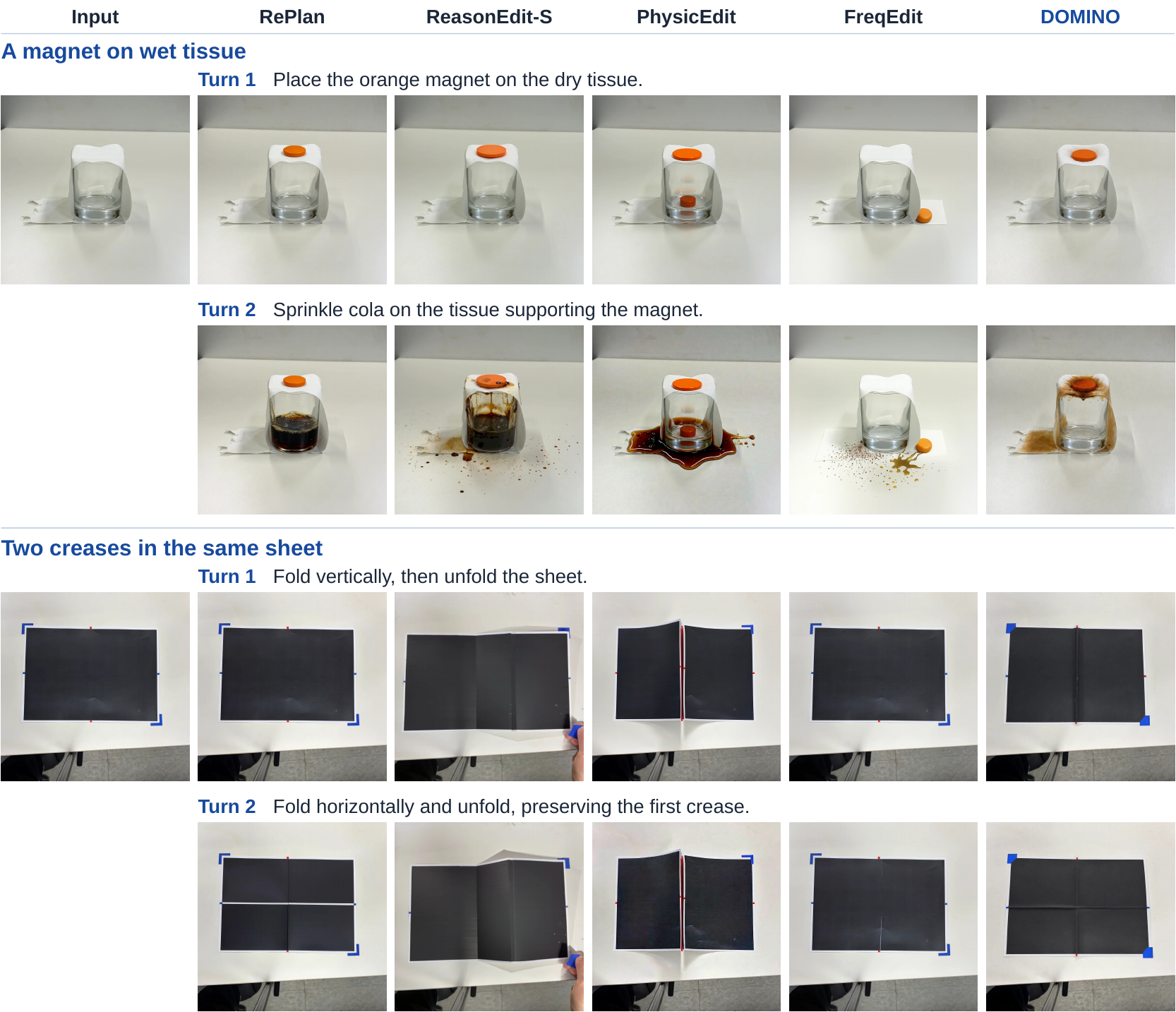}

        \captionsetup{hypcap=false}
        \captionof{figure}{
            \textbf{Wetting a tissue and folding a sheet on \benchmarkname{}.}
            Cola is added to tissue that already supports a magnet.
            A horizontal fold is added to a sheet that already has
            a vertical crease.
        }
        \label{fig:supp_qualitative_results_2}
    \end{minipage}
    \par\vspace{1em}
}]
\raggedbottom

\subsection{Causal Dependencies across Editing Turns}
\label{sec:supp_turn_dependencies}

The two sequences in Fig.~\ref{fig:supp_qualitative_results_2} ask
whether an earlier edit is still visible after a second intervention.
In the first, \modelname{} places the orange magnet on the tissue.
After cola is added, a brown stain extends down the tissue and the
magnet remains on top. FreqEdit places the magnet beside the glass in
the first turn and retains that placement in the second. In the paper
sequence, both creases are visible in the final \modelname{} result.
The PhysicEdit result still shows a pronounced vertical fold.

\begin{figure*}
    \centering
    \includegraphics[width=1\linewidth]{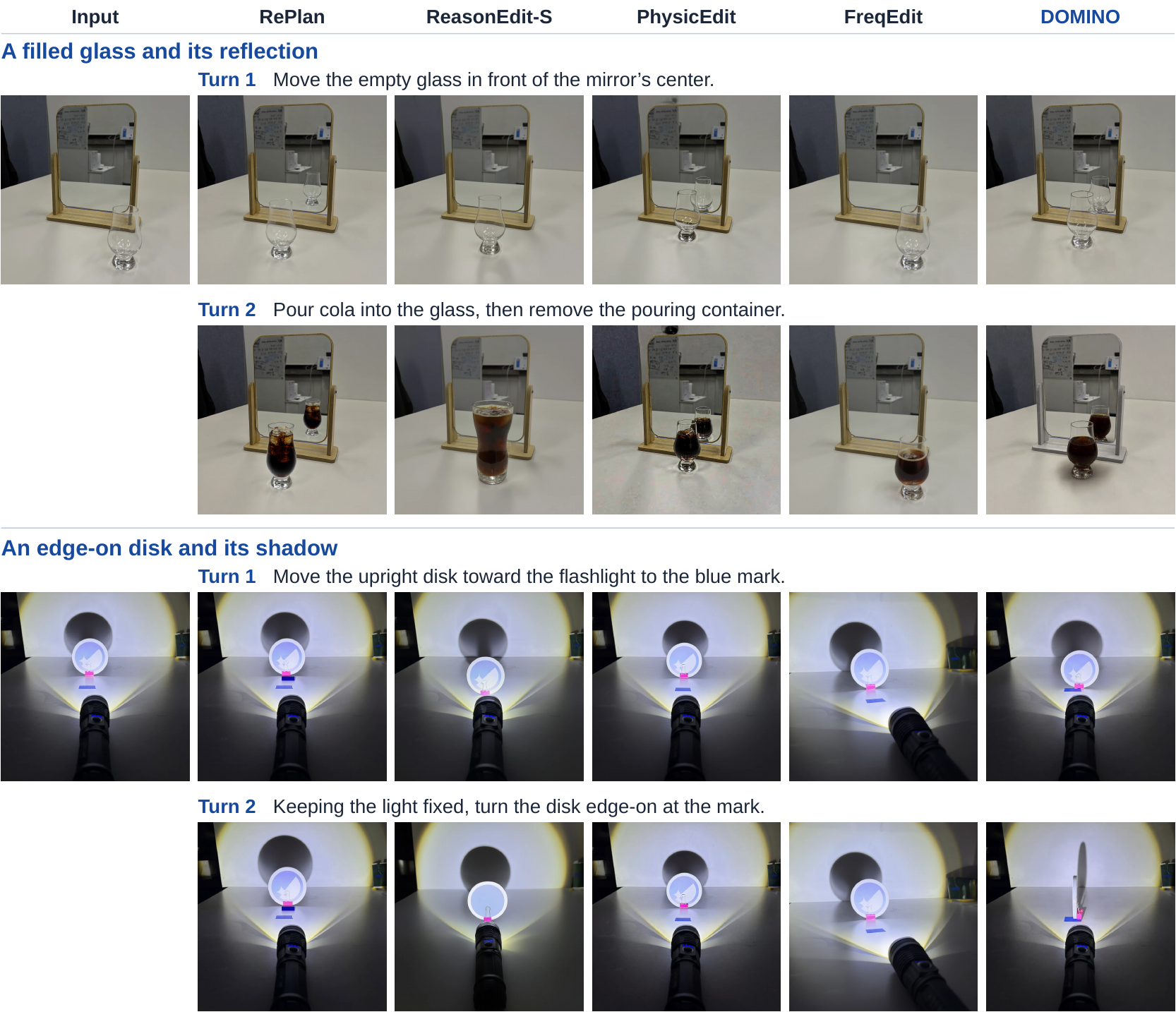}
    \caption{\textbf{Reflections and cast shadows across turns on \benchmarkname{}.}
    Filling the repositioned glass changes its reflection. Turning
    the disk edge-on changes its shadow under the fixed light.}
    \label{fig:supp_turns_mirror_disk}
\end{figure*}

The optical examples in Fig.~\ref{fig:supp_turns_mirror_disk} show why
a later instruction must be interpreted using the current scene. In
the first sequence, moving the glass in front of the mirror establishes
where its reflection should appear. After cola is added, the
\modelname{} result shows the dark liquid in both the glass and its
reflection. In the FreqEdit result, the glass remains to the right of
the mirror. In the second sequence, the disk first moves toward the
flashlight, then turns edge-on at that location. \modelname{} produces
a narrow disk silhouette and cast shadow. The other four methods
retain a circular disk face after the rotation.

\begin{figure*}
    \centering
    \includegraphics[width=1\linewidth]{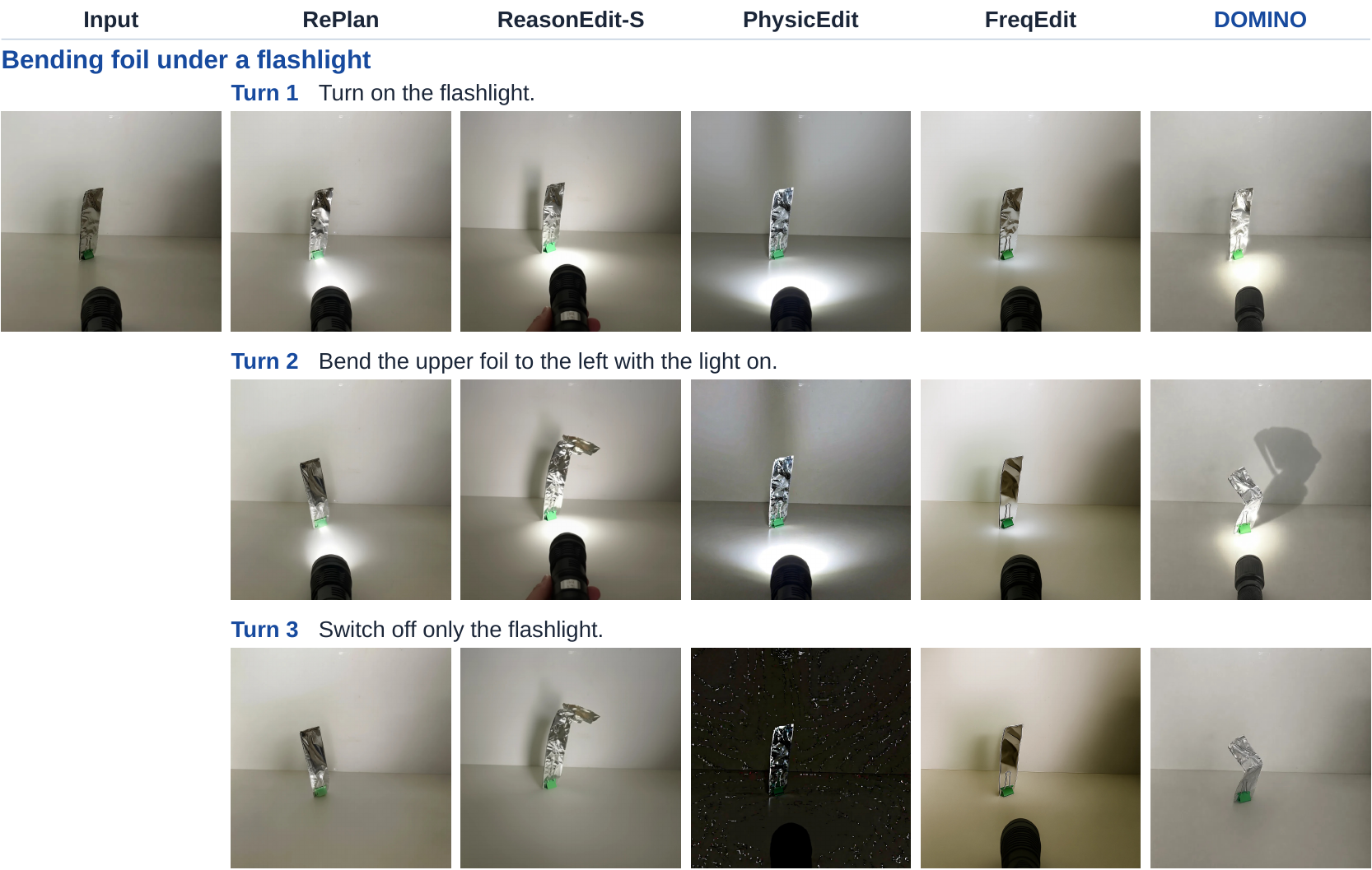}
    \caption{\textbf{Changing illumination while preserving a deformation on \benchmarkname{}.}
    The light is turned on, the foil is bent, and the light is switched
    off. The final scene should retain both the bent foil and the flashlight.}
    \label{fig:supp_turn_dependencies}
\end{figure*}

The flashlight sequence tests an update and a preservation requirement
in the same turn (Fig.~\ref{fig:supp_turn_dependencies}). After the
light comes on, \modelname{} bends the foil and produces a cast shadow.
Switching off the light removes the bright patch and the pronounced
shadow while preserving the bend. The flashlight body also disappears
from the final \modelname{} image, although the instruction requires
it to stay in place.

\section{Limitation}
\label{sec:limit}

\paragraph{Ambiguity in physical outcomes.} We use colored masking tape and other visible landmarks to specify where an action should take place. Even with these spatial cues, an instruction can admit multiple physically plausible outcomes. For example, a tape marker on a water-filled paper cup indicates how far to extend a slit. Water escapes through the extended slit, but the instruction leaves the exact shape of the resulting puddle undetermined. The captured reference therefore records one plausible outcome, and a difference in the puddle’s shape alone does not imply a physical error.

\section{Interactive Demonstrations and Additional Qualitative Examples}
\label{sec:supp_project_page}

The project page, linked below the title on the first page of the main
paper, provides interactive demonstrations of \modelname{}, photographed
examples from \benchmarkname{}, and side-by-side comparisons with other
editing methods.

\paragraph{Method walkthrough.}
The \emph{Method} section follows a single editing turn through
Fig.~\ref{fig:model}. Readers can step through the scene graph built
by the Weaver, the editing plan proposed by the Forecaster, and the
Sentinel's review, then see how CRA associates the reviewed editing
instructions with image regions.

\paragraph{Interactive control.}
The \emph{Controls} section lets readers adjust an image region and
inspect the corresponding recorded edits. For example, dragging the
upper edge of the water region shows how the water level changes
across three settings. Moving the cube region across the top card

\clearpage
\end{document}